%% file: working_note.tex
\documentclass[conference]{edm_template}

\usepackage{booktabs}
\usepackage{cite}
\usepackage[square,sort,comma,numbers]{natbib}
\usepackage{url}            
\usepackage{floatrow, tabularx}
\usepackage[subrefformat=simple,labelformat=simple]{subcaption}
\usepackage[labelfont=bf,textfont=bf]{caption}
\usepackage{mathtools}
\usepackage{subcaption}
\usepackage{multicol}
\usepackage{graphicx}

\DeclarePairedDelimiterX{\infdivx}[2]{(}{)}{%
  #1\;\delimsize\|\;#2%
}

\newcommand{\infdiv}{D_{KL}\infdivx}

\begin{document}

\title {Time-series Insights into the Process of Passing or Failing Online University Courses using Neural-Induced Interpretable Student States}

\numberofauthors{5} 

\author{
\alignauthor 
      Byungsoo Jeon\\
      \affaddr{Carnegie Mellon University}\\
      \affaddr{5000 Forbes Avenue}\\
      \affaddr{Pittsburgh, PA}\\
      \email{byungsoj@cs.cmu.edu}
\alignauthor Eyal Shafran\\
      \affaddr{Western Governors University}\\
      \affaddr{4001 S 700 East}\\
      \affaddr{Salt Lake City, UT}\\
      \email{eyal.shafran@wgu.edu}
\alignauthor Luke Breitfeller\\
      \affaddr{Carnegie Mellon University}\\
      \affaddr{5000 Forbes Avenue}\\
      \affaddr{Pittsburgh, PA}\\
      \email{mbreitfe@cs.cmu.edu}
\and  
\alignauthor Jason Levin\\
      \affaddr{Western Governors University}\\
      \affaddr{4001 S 700 East}\\
      \affaddr{Salt Lake City, UT}\\
      \email{jason.levin@wgu.edu}
\alignauthor Carolyn P. Ros\'e\\
      \affaddr{Carnegie Mellon University}\\
      \affaddr{5000 Forbes Avenue}\\
      \affaddr{Pittsburgh, PA}\\
      \email{cprose@cs.cmu.edu}
}

\maketitle
\begin{abstract}
\input{000abstract}
\end{abstract}


\section{Introduction}
\label{sec:intro}
\input{010introduction}

\section{Related Work}
\label{sec:related_work}
\input{030related_work}

\section{Data}
\label{sec:data}
\input{040proposed_method}

\section{Results}
\label{sec:experiment}
\input{050experiment}

\section{Conclusion}
\label{sec:conclusion}
\input{060conclusion}

\section{Acknowledgement}
\label{sec:acknowledgement}
This work was funded in part by NSF grantIIS 1822831.

\bibliography{working_note}
\bibliographystyle{plain}

\end{document}

%% file: 000abstract.tex
This paper addresses a key challenge in Educational Data Mining, namely to model student behavioral trajectories in order to provide a means for identifying  students most at-risk, with the goal of providing supportive interventions. While many forms of data including clickstream data or data from sensors have been used extensively in time series models for such purposes, in this paper we explore the use of textual data, which is sometimes available in the records of students at large, online universities. We propose a time series model that constructs an evolving student state representation using both clickstream data and a signal extracted from the textual notes recorded by human mentors assigned to each student. We explore how the addition of this textual data improves both the predictive power of student states for the purpose of identifying students at risk for course failure as well as for providing interpretable insights about student course engagement processes.

%% file: 010introduction.tex
With the rapidly-changing landscape of work opportunities causing increased concerns related to unemployment, workers have a greater need to seek further education. Online universities \cite{ashby2004monitoring} like Western Governor's University (WGU) \cite{kinser2002taking} play crucial roles in helping workers achieve their career success by providing a personal and affordable education based on real-world competencies. In such online educational contexts, modeling the population of students at scale is an important challenge, for example, in order to identify students most at-risk and to provide appropriate interventions to improve their chances of earning a degree in a timely fashion. In this respect, a plethora of approaches for clickstream analysis \citep{tang2016deep,fei2015temporal,DBLP:conf/iccse/WangYM17} have been published in the field of Educational Data Mining, which address questions about modeling student course engagement processes.  Some of this work has produced time series models producing predictive signals related to dropout or disengagement, which can then be used as triggers for interventions \cite{rienties2016analytics4action}. While clickstream data is the most readily available, and while some success has been achieved using it for this purpose, its low level indicators provide only glimpses related to student progress, challenges, and affect as we would hope to observe and model them.  In this paper, we explore the extent to which we may achieve richer insights by adding textual data to the foundation provided by clickstream data.

One advantage to modeling student behavior and states from a for-pay platform is that the level of support provided to students is greater than in freely available contexts like Massive Open Online Courses (MOOCs), and this more intensive engagement provides richer data sources that can be leveraged.  In our work, we make use of a new data source provided by the Western Governor's University (WGU) platform, where each student is assigned a human mentor, and the notes from each biweekly encounter between student and mentor are recorded and made part of the time series data available for each student. Thus, even if we do not have access to the full transcript of the interactions between students and their mentors, we can leverage the documentation of provided support in order to enhance the richness and ultimately the interpretability of student states we may induce from other low level behavioral indicators we can extract from traces of learning platform interactions. 

A major thrust of our work has been to develop a technique for leveraging this form of available textual data. We refer to this data as \textit{Mentor's Notes}. In particular, we propose a sequence model to integrate available data traces over time, which we refer to as \textbf{Click2State}, which serves a dual purpose. The first aim is to induce predictive student states, which provide substantial traction towards predicting whether a student is on a path towards passing or failing a course. Another is to provide us with insights into the process of passing or failing a course over time, and in particular leveraging the insights of human mentors whose observations give deeper meaning to the click level behavioral data, which is otherwise impoverished from an interpretability standpoint.

In the remainder of the paper we discuss related work in the fields of Educational Data Mining of large scale online course data to contextualize our specific work.  Next we discuss the specific data we are working with and how it relates to the context in which it was collected. Then we explain from a technical level the modeling approach we are taking in this work. Finally, we present a series of experiments that investigate the following three research questions:

\begin{itemize}
  \item RQ1. How can we leverage contemporary text mining techniques such as Latent Dirichlet Allocation (LDA) \cite{blei2003latent} to extract information and meaning from mentors' notes about the formation of student states across time using a time-series model?
  \item RQ2. To what extent does integrating a representation of topical insights from \textit{Mentor's Notes} improve the ability of a time series neural model to predict whether students are on a path towards passing or failing a course?
  \item RQ3. How can we use insights about student progress in an online course captured using student state representations from our integrated model to understand the process of passing or failing a course on the platform?
\end{itemize}

%% file: 030related_work.tex
\begin{table*}[!ht]
    \centering
        \begin{tabular}{c|c|c|c|c|c|c} 
        \specialrule{.12em}{.1em}{.1em} 
        & \# click & \# focused state & \# keypress & \# mouse move & \# scroll & \# unfocused state \\
        \specialrule{.12em}{.1em}{.1em} 
        Target course & 53 & 61 & 0 & 168 & 904 & 1732\\
        Other courses & 177 & 167 & 0 & 455 & 2301 & 4887\\ 
        Degree plan   & 0 & 0 & 0 & 0 & 0 & 0 \\ 
        Portal        & 21 & 89 & 0 & 263 & 3862 & 2440 \\ 
        Homepage      & 36 & 69 & 0 & 122 & 72 & 1581 \\ 
        \specialrule{.12em}{.1em}{.1em} 
        \end{tabular}
        
	\caption{Example of clickstream data.}
	\label{tab:click}
\end{table*}

\begin{table*}[!ht]
    \begin{tabular}{p{2.6cm}|p{10.5cm}}
        \specialrule{.12em}{.1em}{.1em} 
        Type & Description \\
        \specialrule{.12em}{.1em}{.1em} 
        \# click & The number of mouse clicks. \\
        \# focused state & The number of times a mouse was in a content box. It is incremented by one every time a mouse goes into a content box from somewhere else.\\
        \# keypress & The number of times a keyboard was pressed.\\
        \# mouse move & The number of times a mouse has been moved.\\
        \# scroll & The number of scroll counts.\\
        \# unfocused state & The number of times a mouse was outside a content box.\\
        \specialrule{.12em}{.1em}{.1em} 
    \end{tabular}
    \caption{Description of click types.}
	\label{tab:click_type}
\end{table*}

Past research aiming at enhancing the learning process of students in online universities has focused on providing analytic tools for teachers and administrators. These tools are meant to enhance their ability to offer support and make strategic choices in the administration of learning within the contexts under their care. As just one example, the Analytics4Action Evaluation Framework (A4AEF) model \cite{rienties2016analytics4action}, developed at the Open University (UK) \cite{ashby2004monitoring}, provides a university-wide pipeline allowing its users to leverage learning analytics to enable successful interventions and to gain strategic insights from their trials. One of the most important challenges in implementing such a pipeline is to model the population of students in such a way as to provide both predictive power for triggering interventions and interpretability for ensuring validity and for supporting decision making.  

Some past research has already produced models to identify at-risk students and predict student outcomes specifically in online universities \citep{calvert2014developing, matuga2009self}. For example, Smith et el. \cite{smith2012predictive} proposed models to predict students' course outcomes and to identify factors that led to student success in online university courses. Eagle et al. \cite{eagle2018wgu} presented exploratory models to predict outcomes like high scores on the upcoming tests or overall probability of passing a course, and provided examples of strong indicators of student success in the WGU platform where our work is also situated. However, this past work has focused mainly on predictive modeling of student outcomes, whereas our work pursues both predictive power and interpretability. 

While much work in the field of Educational Data Mining explores time series modeling and induction of student state representations from open online platforms such as Massive Open Online Courses (MOOCs) or Intelligent Tutoring Systems, far less has been published from large, online universities such as WGU, which offer complementary insights to the field. Student states are triggered by students' interaction with university resources, their progress through course milestones, test outcomes, affect-inducing experiences, and so on. Affect signals in particular have been utilized by many researchers as the basis for induced student states, as this rich source of insight into student experiences has been proven to correlate with several indicators of student accomplishments \citep{rodrigo2009affective, pardos2014affective, craig2004affect}. Researchers have investigated affect and developed corresponding detectors using sensors, field observation, and self-reported affect. These detectors capture students' affective signals from vocal patterns \citep{calvo2010affect, nicolaou2011continuous}, posture \cite{d2007mind}, facial expressions \citep{bosch2014s, nicolaou2011continuous}, interaction with the platform \citep{Botelho2018StudyingAD, botelho2017improving, heffernan2014assistments}, and physiological cues \citep{liu2005empirical, calvo2010affect}. Although these signals provide rich insights, the requisite data is sometimes expensive or even impractical to obtain, even on for-pay platforms such as WGU, where we conduct our research.

The bulk of existing work using sequence modeling to induce student states has focused on the data that is most readily available, specifically, clickstream data. For example, Tang et al. \cite{tang2016deep} have constructed a model to predict a set of student actions with long short-term memory (LSTM) \cite{hochreiter1997long} on student clickstream data from a BerkeleyX MOOC, though the basic LSTM was unable to match the baseline of defaulting to the majority class for samples of student actions. Fei et al. \cite{fei2015temporal} proposed a sequence model to predict dropout based on clickstream data using deep learning models such as recurrent neural networks (RNNs) and LSTMs, with more success. Wang et al. \cite{DBLP:conf/iccse/WangYM17} also built a deep neural network architecture using a combination of convolutional neural network (CNN) \cite{krizhevsky2012imagenet} and RNN for dropout prediction from clickstream data. Though these models have achieved differing success at their predictive tasks, a shortcoming shared by all of these models is the lack of interpretability in the induced student state representations.

Some prior work has nevertheless attempted to construct cognitively meaningful representations, such as representations that can be constructed through summarization of raw video clickstream data \citep{SinhaJLD14, Sinha2014}. Sinha et al. \cite{SinhaJLD14} attempted to construct cognitive video watching states to explain the dynamic process of cognition involved in MOOC video clickstream interaction. Building on this work, Sinha et al. \cite{Sinha2014} explored the combined representations of video clickstream behavior and discussion forum footprint to provide insights about student engagement process in MOOCs. Similar to our own work, their work extracting these abstract feature representations from the raw clickstream data aims (1) to obtain noise-resistant and interpretable features and (2) to transform the unstructured raw clickstream data to structured data appropriate as input to existing statistical or machine learning models. However, this published work does not model the temporal patterns of such cognitively meaningful features such as we perform in this paper. Our work extends previous studies by proposing a model that enriches temporal signals from clickstream data using the textual mentor's notes in order to provide a means for interpreting student state representations as they evolve over time within courses.

%% file: 040proposed_method.tex
Our study is based on data collected by Western Governor's University (WGU), an online educational platform \footnote{\scriptsize\url{https://www.wgu.edu/}}. WGU is an online school with career-focused bachelor's and master's degrees---in teaching, nursing, IT, and business---designed to allow working professionals the opportunity to fit an online university education into their busy lives. Students in WGU earn a competency-based degree by proving knowledge through regular assessments \cite{johnstone2005competency}, which facilitates self-paced learning based on their prior experience. 

\begin{figure}[H]
	\centering
    \includegraphics[width=0.7\linewidth]{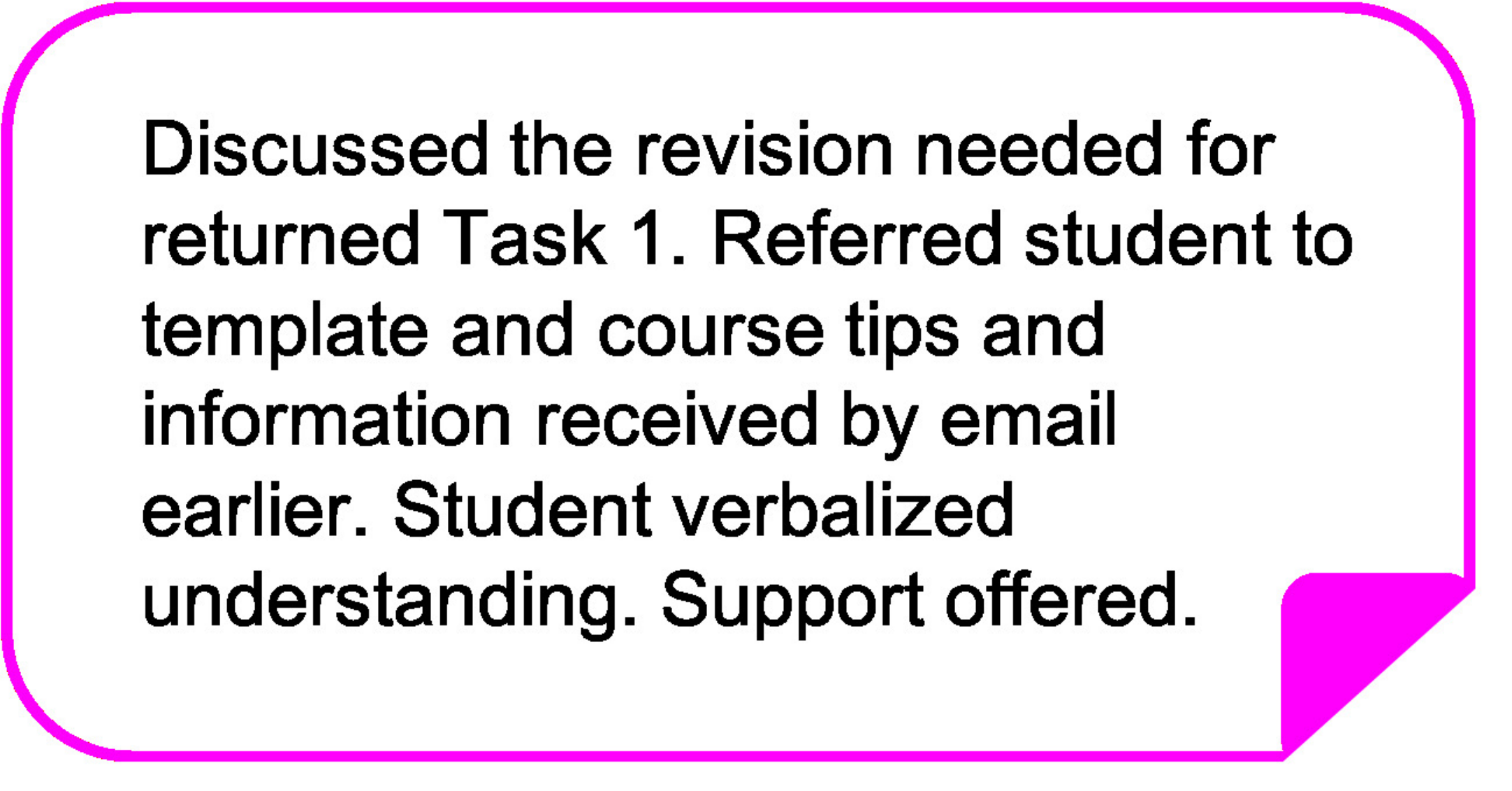}
    \caption{An example of mentor's notes.}
    \label{fig:mnote}
\end{figure}

To support self-paced learning, students in WGU are assigned to a program mentor (PM). The PM is in charge of evaluating a student's progress through their degree and helping to manage obstacles the student faces. A PM and a student generally have bi-weekly live calls, but this may vary depending on the student's needs and schedule. Each PM writes down a summary of what was discussed, which we refer to as a mentor's note. An example is given in Figure \ref{fig:mnote}. As in the example, mentor's notes describe the status and progress of the student and what types of support was offered or what suggestions were made during the call. This information can provide meaningful cues to infer student states over time.

In our modeling work we aim both for predictive power and interpretability, thus it is an important part of our work to build models that induce an interpretable student state representation. In this work we specifically investigate how the use of the mentor's note data along side the more frequently used clickstream data might enable that important goal. Clickstream data in WGU also provides us with information on how active students are and where in the WGU platform they spend their time. We collect clickstream data from four different types of web pages in the WGU platform: course, degree plan, homepage, and portal. The course web pages cover all pages related to courses in WGU. Degree plan represents a dashboard where students check their progress toward a degree. Homepage is the main page that shows students' progress in each course and allows access to all provided WGU features. Portal covers any other pages for student support including technical and financial assistance. 

An example of the clickstream data can be seen in Table \ref{tab:click}. Each row represents one of five different click sources: target course page, other course page, degree plan page, portal page, and homepage. We divide the course pages into "target course" and "other course". Each column represents one of different six click types: click count, focus state count, keypress count, mousemove count, scroll count, and unfocused state count. The values in the table represent the weekly count of different type of clicks from each different source. The description of these click types are in Table \ref{tab:click_type}. 

\begin{table}[htb]
    \caption{Data Statistics \label{tab:data_stat}}
    \resizebox{\columnwidth}{!}{%
        \begin{tabular}{l|c|c}
        \specialrule{.12em}{.1em}{.1em} 
         & HA & CA\\
        \specialrule{.12em}{.1em}{.1em} 
        \# of students & 6,041 & 4,062\\
        Length of a term  & 25 weeks & 25 weeks\\
        Avg prior units  & 62 $\pm$ 39 & 11 $\pm$ 23\\
        Fail rate  & 0.185 & 0.509 \\
        Avg \# of notes per student  & 10.9 $\pm$ 5.7 & 11.0 $\pm$ 5.8\\
        Avg length of notes (chars)  & 198 $\pm$ 47 & 194 $\pm$ 55\\
        \specialrule{.12em}{.1em}{.1em} 
        \end{tabular}
    }
\end{table}
\vfill\null
\columnbreak

For this paper, we have collected the mentor's notes and clickstream data from two courses conducted in 2017: Health Assessment (HA) and College Algebra (CA). We choose these two courses because they are popular among students and represent different levels of overall difficulty. Table \ref{tab:data_stat} shows the statistics for the dataset. ``Average prior units" is the average number of units students transferred to WGU from prior education when they started the degree, and functions as a proxy for the level of student's prior knowledge. We split the dataset for each course into a training set (80\%), a validation set (10\%), and a test set (10\%). For training, in order to avoid a tendency for trained models to over-predict the majority class, we have resampled the training set so that both the pass state and the fail state are represented equally. 


\begin{figure*}[!ht]
	\centering
	\includegraphics[width=.65\linewidth]{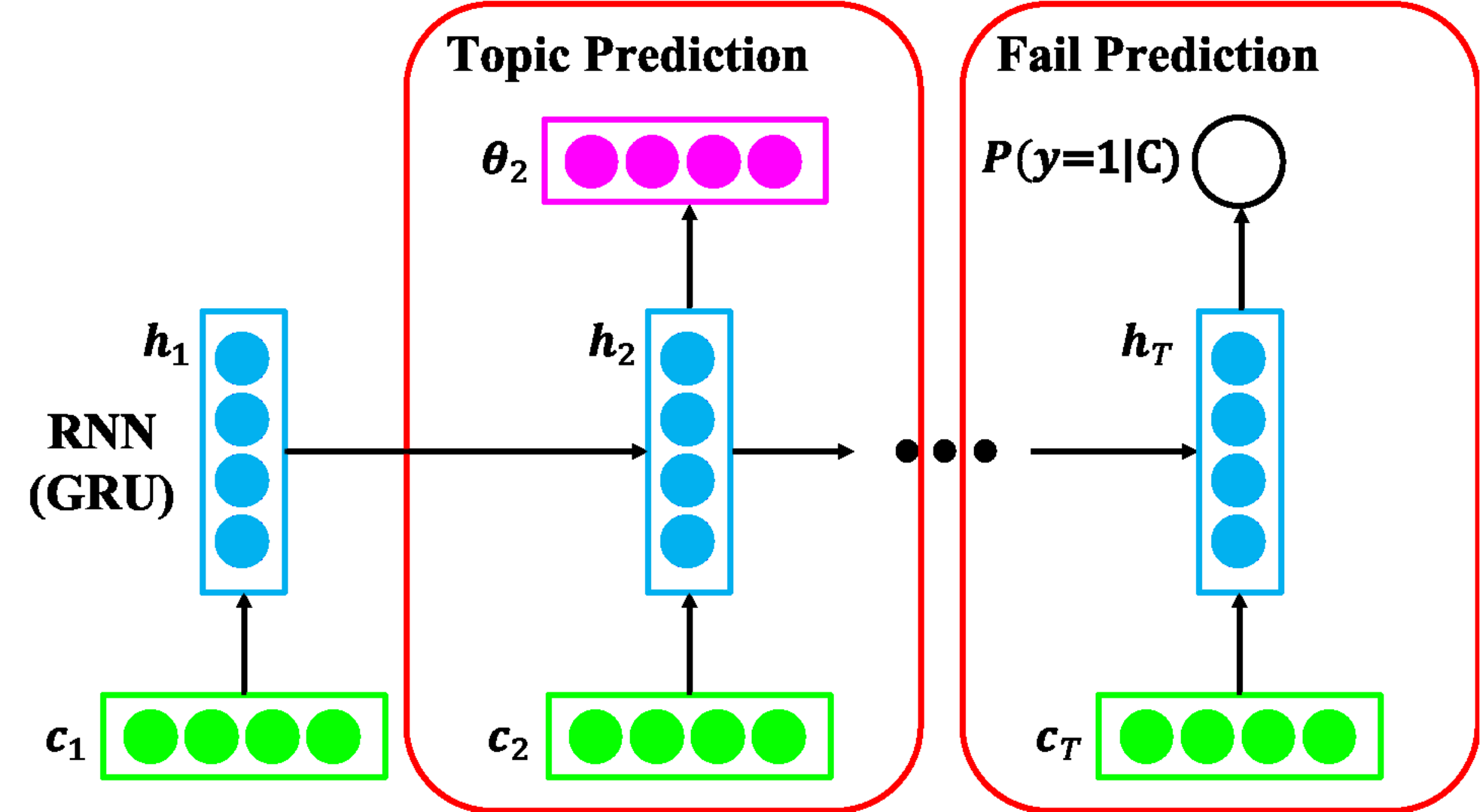}
	\caption{Architecture of Click2State Model.}
	\label{fig:method}
\end{figure*}

\section{Proposed Method}
\label{sec:proposed_method}

As we have stated above, in our modeling work, we propose a sequence model, \textbf{Click2State}, with two primary purposes. The first is to form a student state representation that will allow us to better identify students at risk of failing a course than a baseline model that does not make use of rich textual data. The second is to provide us with a means to interpret the meaning of a student state representation.

Figure~\ref{fig:method} provides a schematic overview of our proposed model.  Note that it is first and foremost a sequence model that predicts whether a student will pass or fail a course based on an interpretable student state that evolves from week to week as each week's clickstream data is input to the recurrent neural model. A summary of the content of the mentor's note for a week is constructed using a popular topic modeling technique, specifically Latent Dirichlet Allocation (LDA). In the full model, an intermittent task to predict the topic distribution extracted from the mentor's notes associated with a time point is introduced. The goal is to use this secondary task to both improve the predictive power of the induced student states over the baseline as well as to enhance the interpretability of the state respresentation. Below we evaluate the extent to which both of these objectives are met in this model.

When training a neural model, inputs are presented to the network, activation is propagated forward, then an error is computed at the output, and the error is propagated backwards through the network.  As the error is propagated backwards, adjustments are made to the weights in order to reduce the occurrence of such errors.  In all of our experiments, the only input at each time step is a representation of the clickstream data from the time step. We have two predictive tasks, namely pass prediction and topic prediction (from mentor's notes). We propagate errors for pass prediction only after the whole sequence of inputs has been provided to the network. Error for topic prediction of mentor's notes are propagated after each time step where mentor's notes are provided with the clickstream data.



\newcommand{\BCE}{\it{BCE}}
\newcommand{\KLD}{\it{KLD}}
\newcommand{\Ndata}{N^{D}}

\textbf{Feature Vector Design}
We train our model using clickstream feature vectors (as input) and topic distribution vectors (as output for the topic prediction task). We design the clickstream feature vector to include both an encoding of click behavior of students from a time period as well as a control variable that represents the prior knowledge of students as estimated by the number of units they were able to transfer in from their higher education experience prior to starting a degree at WGU. The full clickstream feature vector contains thirty weekly counts for each different type and source of click, in addition to the single control variable just mentioned, which is the number of transferred units. We use min-max normalization to scale the values between 0 and 1 in preparation for training. To extract a topic distribution vector for each mentor's note, we run Latent Dirichlet Allocation (LDA)\cite{blei2003latent} over the whole set of mentor's notes from the entire training dataset.

\textbf{Formal Definition of the Model} 
Now we formally specify the model. Denote the student's clickstream features by $C = (c_1, c_2, ..., c_T)$, where $c_t$ is the clickstream feature vector of $t$th week, and $T$ is the number of weeks for the term. The clickstream feature vectors are encoded via Gated Recurrent Units (GRU) \cite{cho2014properties}, which are variants of the Recurrent Neural Network (RNN). Each time step is a week, $t$.  Thus, at each time step $t$, this network constructs a hidden state of the student for the $t$th week, $h_t \in \mathbb{R}^{H}$, where $H$ is the dimensionality of the hidden state representation. We consider $h_t$ as the student state representation associated with the $t$th week. Based on the generated student state representation from RNN ($h_t$), our model is trained to predict a topic distribution of a mentor's note.  As mentioned above, the model is only trained to predict the topic distribution  $\hat{\theta_t}$ for every week $t$ where a student has a mentor's note in the data. But at test time, a topic distribution can be predicted for every student state since the representation of student state is always within the same vector space. Similarly, though we only propagate error for the pass or fail prediction task after a series of weeks of data for a student have been input to the RNN, the pass or fail prediction can be made from the student state representation at any time step.

\textbf{Topic Prediction}
Given the generated hidden states from RNN ($h_t$) for the $t$th week, the model estimates the true topic distribution ($\theta_t \in \mathbb{R}^{N_t}$) of a mentor's note on $t$th week where $N_t$ is the number of topics. The estimated topic distribution ($\hat{\theta_t} \in \mathbb{R}^{N_t}$) is computed by taking $h_t$ as an input of one fully connected layer (weight matrix: $W_{\theta}$) whose output dimensionality is $N_t$ followed by a softmax layer.
\begin{align*}
 \hat{\theta_t} =& \  Softmax(W_{\theta}h_t)
\end{align*}
In training time, the loss function we use for error propagation and adjustment of weights is calculated by means of the Kullback-Leibler divergence loss between $\hat{\theta_t}$ and $\theta_t$.

\textbf{Fail Prediction}
As data from a student's participation in a course is fed into the RNN week by week, the model estimates the probability of the student failing that course ($P(y=1 | C)$) at the last timestep $T$. The estimated probability is computed by taking $h_t$ as an input of one fully connected layer (weight matrix: $W_{y}$) whose output dimensionality is one followed by a sigmoid layer.
\begin{align*}
 P(y=1 | C) =& \  Sigmoid(W_{y}h_t)
\end{align*}
In training time, the loss function we use for error propagation and adjustment of weights for fail prediction is the computed binary cross entropy loss with $P(y=1 | C)$ and a true label for $n$th student, $y_n$.

\textbf{Loss}
The loss function is composed of KL divergence loss for the topic prediction and binary cross-entropy loss for the fail prediction. Assume there are a total of $N$ students. The KL divergence loss of topic distribution of the mentor's note for $n$th student at time $t$ is defined as:
\begin{align*}
\KLD_{n,t} =& \ \infdiv{\theta_{n,t}}{\hat{\theta}_{n,t}},
\end{align*}
where $\theta_{n,t}$ and $\hat{\theta}_{n,t}$ are the true and estimated topic distribution of the mentor's note at time $t$ for $n$th student.

The binary cross-entropy of the $n$th student measures the similarity between the predicted $P(y=1 | C)$ and the true $y$ as:
\begin{align*}
\BCE_{n} =& - y_n \log P_\Theta(y_n=1 | C) \\
&- (1 - y_n) \log (1 - P_\Theta(y_n=1 | C)),
\end{align*}
where $y_n$ is the true $y$ ($\in \{0, 1\}$) of the $n$th student and $P_\Theta$ is the probability of the fail predicted by our model with parameters $\Theta$. 

Assume that there are a total of $N_{n}$ mentor's notes for $n$th student. Combining the two losses, our final loss is
\begin{align*}
\frac{1}{N} \sum_{n = 1}^{N} \lbrack \lambda \BCE_{n} + (1-\lambda)\frac{1}{N_{n}}\sum_{t = t_{n,1}}^{t_{n,N_n}}\KLD_{n,t}\rbrack,
\end{align*}
where $t_{n,i}$ is the timestep when $n$th student has $i$th mentor's note, and $\lambda$ is the rescaling weight to balance the contribution of two different loss functions.


\newcolumntype{L}[1]{>{\raggedright\let\newline\\\arraybackslash\hspace{0pt}}m{#1}}
\newcolumntype{C}[1]{>{\centering\let\newline\\\arraybackslash\hspace{0pt}}m{#1}}
\newcolumntype{R}[1]{>{\raggedleft\let\newline\\\arraybackslash\hspace{0pt}}m{#1}}

\begin{table*}[!ht]
    \centering
        \begin{tabular}{C{3cm}|L{4.8cm}|L{8cm}} 
            \specialrule{.12em}{.1em}{.1em} 
            Topic & \multicolumn{1}{c|}{Topical Words} & \multicolumn{1}{c}{Topical Text}\\
            \specialrule{.12em}{.1em}{.1em} 
            T1. Revision & task, submit, revise, discuss, equate, complete, need, write, practice, paper & The ST and I discussed his \textbf{Task} 3 \textbf{revisions} after he made some corrections. The ST still needs to \textbf{revise} the \textbf{task} based on the evaluator's comments. He plans to do more \textbf{revisions} that align with the \textbf{task} rubric and \textbf{submit} the \textbf{task} soon.\\
            \hline
            T2. Question & student, question, call, email, send, course, discuss, appoint, speak, assist & \textbf{Student} \textbf{emailed} for help with getting started.  CM called to offer support.  \textbf{Student} could not talk for long.  CM \textbf{emailed} welcome letter and scheduling link and encouraged for student to make an appointment \\
            \hline
            T3. Assessment & week, goal, today, schedule, pass, take, exam, final, work, talk & C278: \textbf{took} and did not \textbf{pass} preassessment, did not \textbf{take} \textbf{final}.  NNP C713: \textbf{took} and did not \textbf{pass} the preassessment. \textbf{Passed} LMC1 PA with a 65 on 02/27. LMC1 \textbf{exam} \textbf{scheduled} for 02/27 \\
            \hline
            T4. Review for exam & student, review, assess, plan, study, attempt, discuss, complete, take, report & Student scheduled appointment to \textbf{review} for first OA \textbf{attempt} but had \textbf{taken} and not passed the \textbf{attempt} by the time of the appointment.\\
            \hline
            T5. Term plan & student, discuss, course, complete, engage, college, term, plan, pass, progress & \textbf{Discussed} final \textbf{term} \textbf{courses}. \textbf{Discussed} starting C229 and working through hours and then working through C349 \textbf{course}.\\
            \hline
            T6. Course progress & goal, course, progress, current, complete, previous, work, date, pass, module & Current \textbf{course}: C349 Previous goal: \textbf{completed} \textbf{modules} 1-3 and engage shadow health by next appt date \textbf{Progress} toward goal: Yes New Goal: shadow health \textbf{completed} and engaged in video assessment\\
            \hline
            T7. Term progress & term, course, complete, date, goal, week, progress, current, leave, remain & Date:  8/22/17 \textbf{Term} Ends: 5 \textbf{weeks} OTP \textbf{Progress}:  5/14 cu \textbf{completed} Engaged \textbf{Course}: C785 \textbf{Goal} \textbf{Progress}: did not pass PA \\
            \hline
            T8. Time constraint & work, week, lesson, complete, go, progress, plan, finish, time, goal  & NNP stated he was not able to make forward \textbf{progress} in course related to personal situation and \textbf{time} constraints from an unexpected event.\\
            \hline
            T9. Goal setting & goal, week, work, complete, task, progress, pass, accomplish, finish, contact & Previous \textbf{goal}: \textbf{finish} shadow health, \textbf{finish} and submit video by next call, start c228 next \textbf{Progress}/concerns: states \textbf{working} on c349 SH, discussed deadlines  \textbf{Goal}: \textbf{finish} shadow health \\
            \specialrule{.12em}{.1em}{.1em} 
        \end{tabular}
    
    \caption{LDA Topics Learned From Mentor's Notes}
    \label{tab:topic}
\end{table*}



%% file: 050experiment.tex
In this section, we answer our aforementioned research questions one by one. First we describe topics learned from mentor's notes and how they may relate to student states. Then we illustrate experiment results to evaluate our Click2State model, along with experimental settings. We conclude by providing methods of extracting insights from these learned student state representations related to the process of passing or failing a course over time.

\textbf{RQ1. What types of information about student states can we extract from mentor's notes using LDA?}
Mentor's notes are summaries of bi-weekly live calls where program mentors (PM) interact with students to provide advice and support. On this bi-weekly call, PMs mostly check students' progress and help them to establish appropriate study plans to achieve their goals for the term. PMs also diagnose students' issues and developmental needs to better provide struggling students with tailored instruction and support. 

With this in mind, we answer the question of how student state information may be extracted from mentor's notes through application of LDA to the notes. This provides us with topics we can easily describe and interpret to deduce overall student states. In this section, we illustrate the insights this LDA approach yields. We set the number of topics to ten to maximize the interpretability of the results. Table \ref{tab:topic} shows the learned topics with manually assigned labels, topical words, and text. Topical words are the top ten words with the highest probability of appearing within each learned topic, and are presented in decreasing order of likelihood. The topical text column contains an example snippet from one of top ten mentor's notes for each topic with the highest topic probability. Frequently in topic models induced by LDA, not all of the topics are strongly thematic.  Since many words in a text are non-thematic, and since every word needs to be assigned to a topic by the model, one or more learned topics is not coherent enough to interpret.  Thus, we exclude from our interpretation the one topic that was incoherent out of the 10 learned topics, and thus we have nine topics in Table \ref{tab:topic}.

Note that there are two topics related to student progress, course progress (T6) and term progress (T7). Course progress (T6) focuses on progress towards modules in a particular course, along with past and present goals about the course itself. Term progress (T7) emphasizes the number of course units that have been achieved, course units that remain, and goals about courseload within in a term. There is a clear utility to these topics as an interpretation tool for regulation of the student's process moving through the curriculum--if a student hits an impasse in their studies, mentor's notes are expected to focus on what challenges the student experienced and what was discussed to address these challenges. 

We also find two topics directly associated with plans and goals, which are term plan (T5) and goal setting (T9). Term plan (T5) includes discussions about plans for a term, such as course selection and long-term degree planning. Goal setting (T9) is similar in focus to course progress (T6), but is not constrained to a single course. As with the previous topics, these reflect serious investment from the student and make useful cues for favorable progress.

The remaining six topics provide insight on specific issues and circumstances a student may be facing at a particular time, and which may end up impacting their overall progress. In revision (T1), we discover students seeking feedback on revisions, suggesting significant engagement with the platform. In question (T2), students ask for tips on using WGU platforms, course logistics, and how to succeed in a given course. In time constraint (T8), students point out time constraints in their daily life to explain why goals were not met. The time constraint (T8) topic may explain abnormal absence or dropout. Assessment (T3) contains the result or plan of assessments and review for exam (T4) includes progress or plans of review for exam preparation.


\begin{figure*}[!ht]
  \begin{subfigure}[t]{0.32\columnwidth}
    \includegraphics[width=\linewidth]{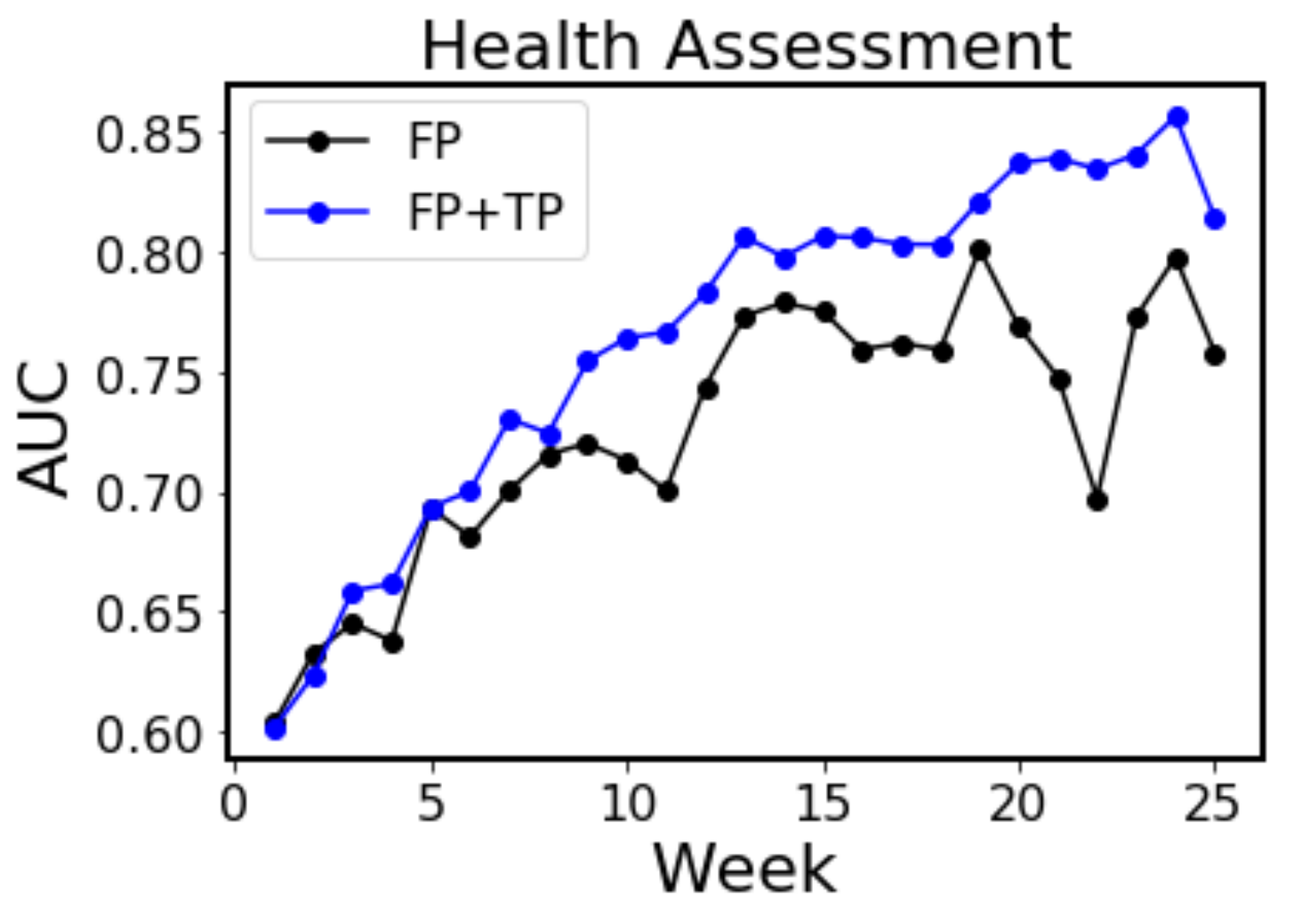}
    \caption{Fail Prediction Performance on Health Assessment by Weeks}
    \label{fig:ha_performance}
  \end{subfigure}
  \hfill 
  \begin{subfigure}[t]{0.32\columnwidth}
    \includegraphics[width=\linewidth]{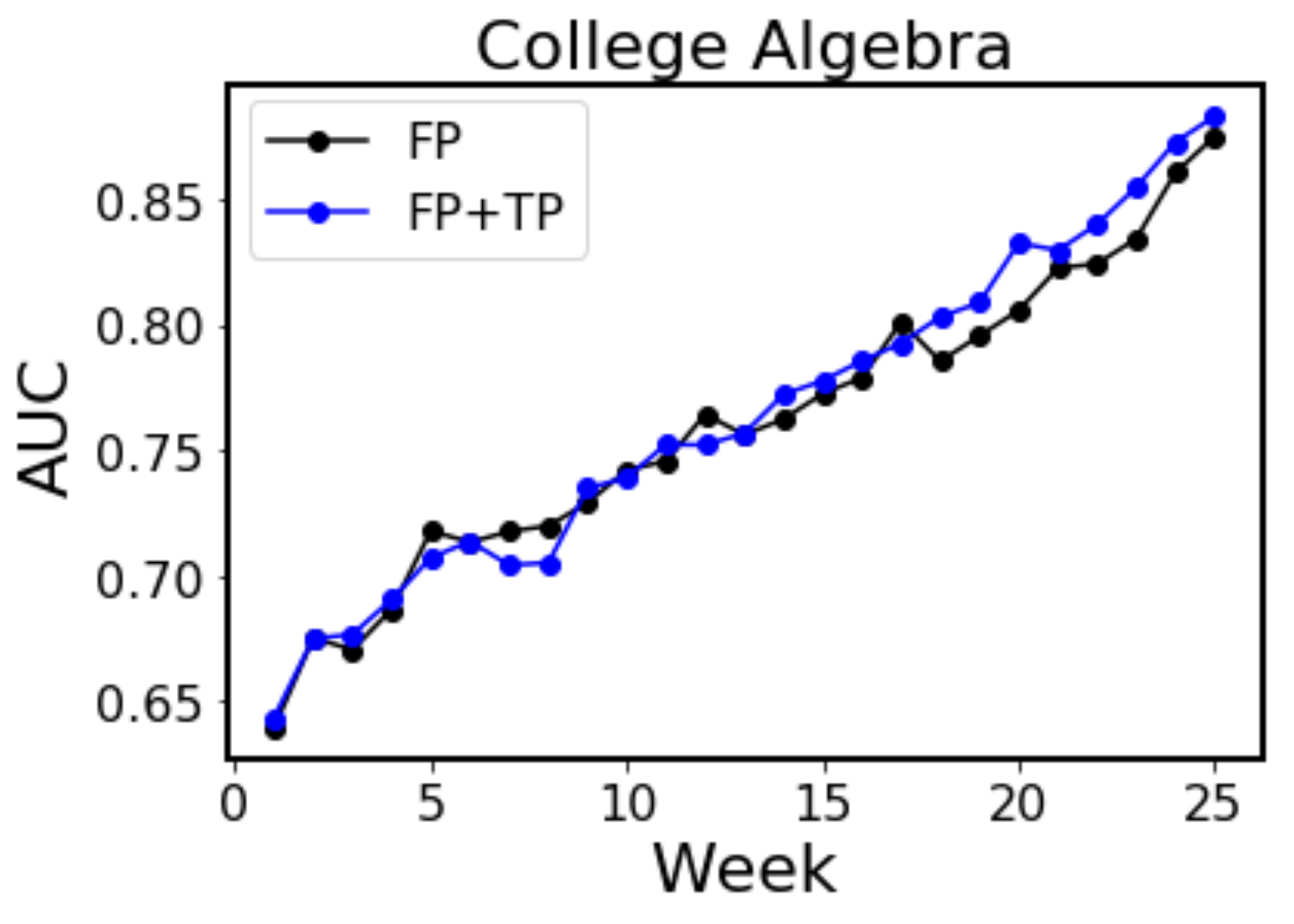}
    \caption{Fail Prediction Performance on College Algebra by Weeks}
    \label{fig:ca_performance}
  \end{subfigure}
    \hfill 
  \begin{subfigure}[t]{0.32\columnwidth}
    \includegraphics[width=\linewidth]{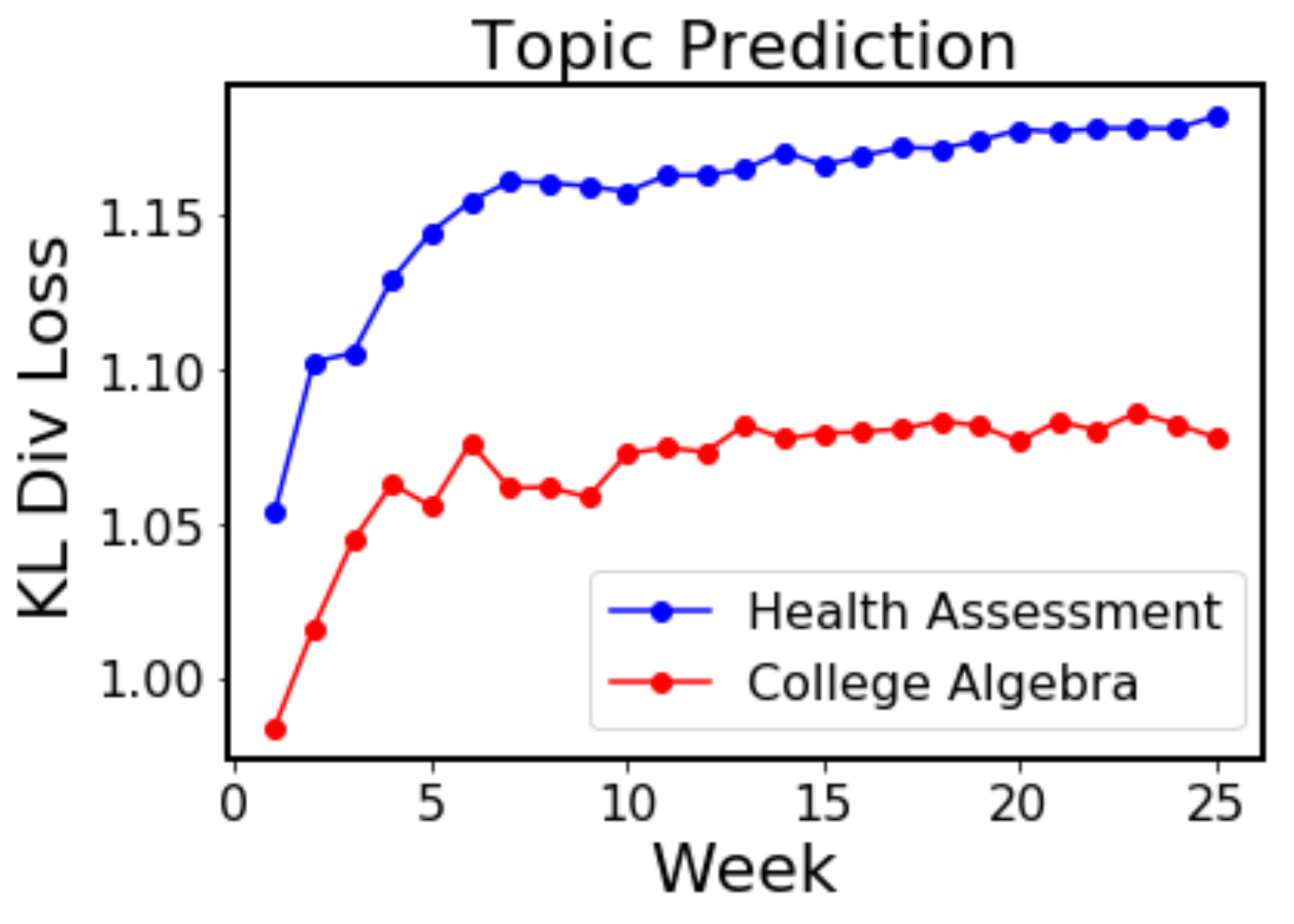}
    \caption{Topic Prediction Performance on Both Courses by Weeks}
    \label{fig:tp_performance}
  \end{subfigure}
  \caption{Performance of Fail Prediction and Topic Prediction by Weeks}
  \label{fig:model_performance}
\end{figure*}
\vfill\null
\columnbreak

\textbf{RQ2. Does the task of topic prediction construct better student state representation than our baseline, as evaluated by the ability to predict student failure?} One method of evaluating the quality of a state representation is by measuring its predictive power on a task, such as the task of predicting whether a student will fail a course. We measure the predictive power of learned student state representation from our model and compare with that of our baseline, which shares the same neural architecture but is not trained on the extra task of topic prediction. The specific predictive task is to determine whether a student fails a course within a given term given a sequence of weeks of student clickstream data. For this analysis, we trained separate models to make a prediction after a set number of weeks so that we could evaluate the difference in predictive power of student states depending on how many weeks worth of data were used in the prediction. The classifier associated with the i-th week is trained on the clickstream data of all students up until the i-th week in the training set, to predict failure as determined at the end of the term. During the training step, we mark the clickstream sequences of the students who failed in the given term as positive instances, and those of students who did not as negative instances.

We evaluate our model and baseline using data from two WGU courses: Health Assessment and College Algebra. As our evaluation metric, we use the AUC score (Area Under the Receiver Operating Characteristic Curve), which measures the probability of a positive instance being classified as more likely to fail a course than a negative instance. For clickstream feature encoding, Gated Recurrent Units (GRU) \cite{cho2014properties} with hidden state sizes 20, 40, 60, 80, and 100 are explored. Optimization is performed using Adam \cite{kingma2014adam} with the initial learning rate 0.001 without weight decay. The rescaling weight ($\lambda$) is 0.5, and the minibatch size is 1. 

Figure \ref{fig:ha_performance} shows the AUC scores across time steps for the Health Assessment course while Figure \ref{fig:ca_performance} shows the AUC scores across time steps for College Algebra. For the Health Assessment course, our model achieves a statistically significant improvement (p-value < 0.05) in performance over the baseline model after the 5th week. For College Algebra, our model achieves a statistically significant improvement after the 17th week. This difference in model performance between the Health Assessment and College Algebra courses suggests the result from College Algebra-specific topic data adds limited predictive power to the model. It is possible the clickstream data of students taking College Algebra already contains enough information about whether a student is going to fail, a conclusion supported by the fact that AUC scores of the baseline model for  College Algebra are always better across time steps than those for Health Assessment. 

We also plot the KL Divergence loss of our model across time steps to determine how well our model is predicting the topic distribution of each mentor's note. The KL Divergence is a typical performance measure when predicting a discrete probability distribution, which measures the entropy increase due to the use of an approximation to the true distribution rather than the true distribution itself. Figure \ref{fig:tp_performance} shows the minimum KL Divergence loss of the topic prediction task for both courses at each week. Though we determined that adding this task improves the fail prediction task, results on this task specifically are not impressive, which demonstrates the relative difficulty of predicting mentor's notes from click data. The KL Divergence loss increases as the number of weeks increases. A possible explanation is that topics become more diverse as the weeks pass and therefore become harder to predict. In addition, the KL Divergence loss is lower in College Algebra than in Health Assessment, which could be explained by students in College Algebra having less variance in the topics of focus in their mentors' notes.

\begin{figure*}[!ht]
  \begin{subfigure}[t]{0.3\columnwidth}
    \includegraphics[width=\linewidth]{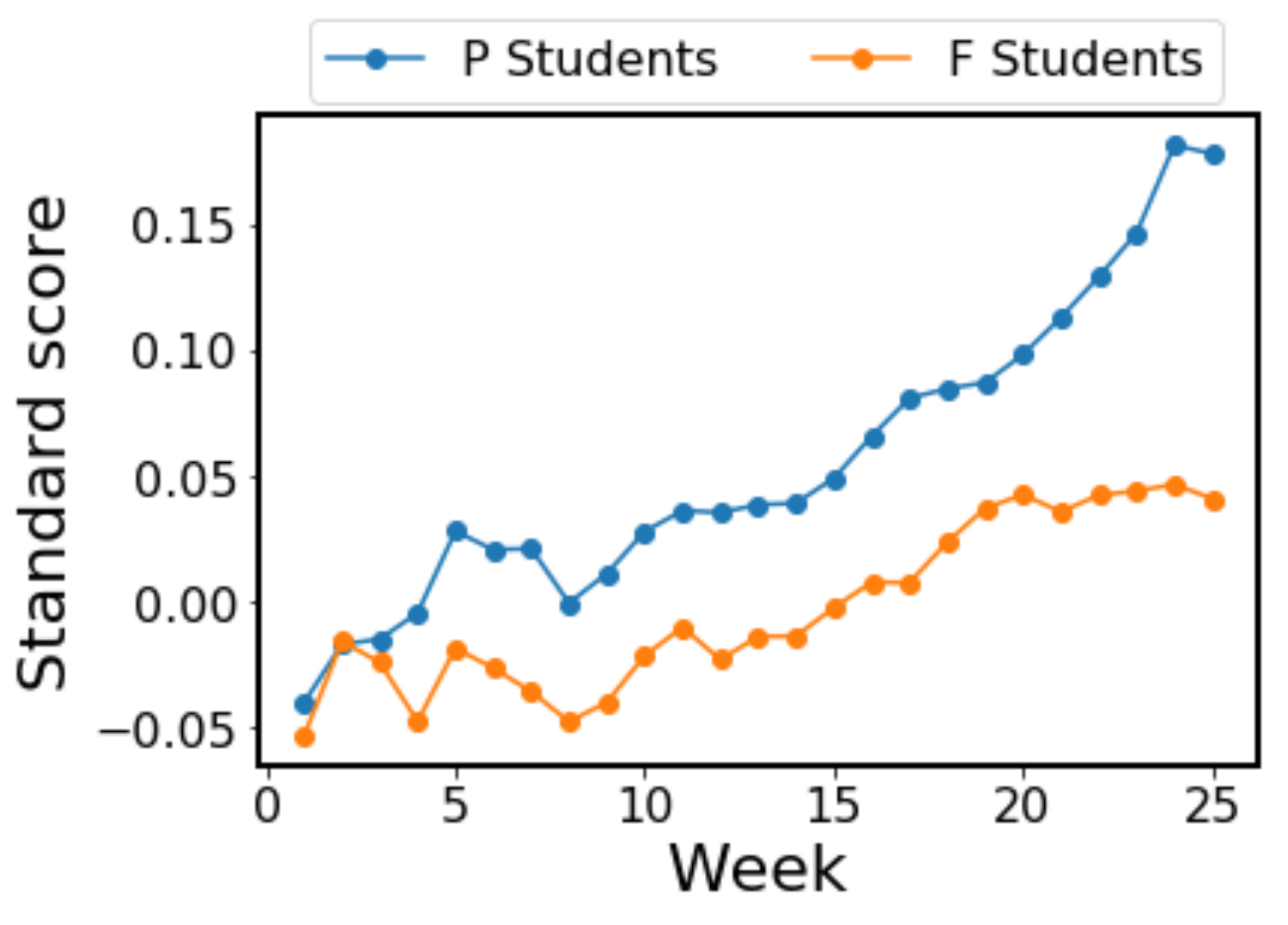}
    \caption{Topic "Revision"}
    \label{fig:case_complete}
  \end{subfigure}
  \hfill 
  \begin{subfigure}[t]{0.3\columnwidth}
    \includegraphics[width=\linewidth]{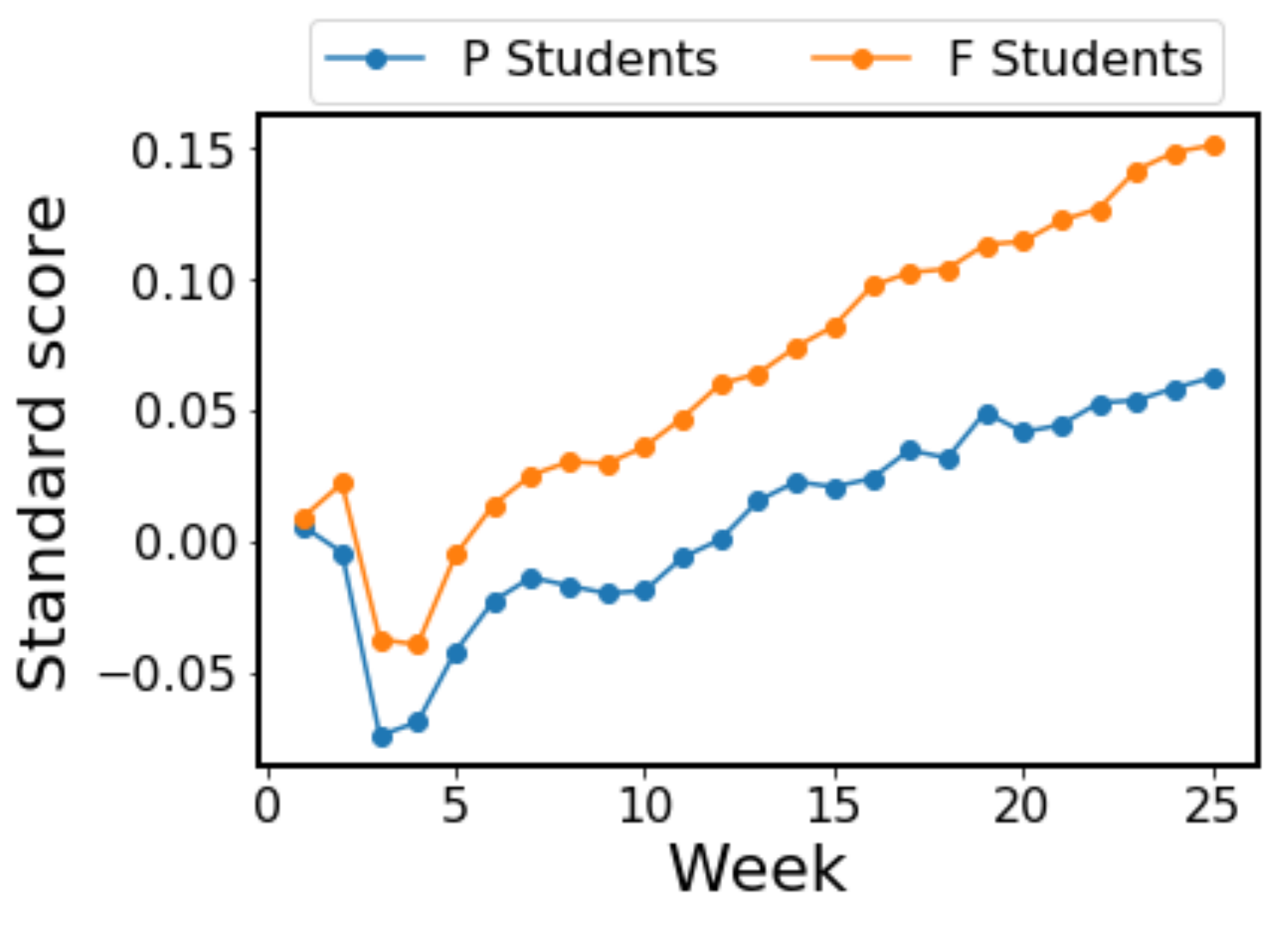}
    \caption{Topic "Question"}
    \label{fig:term_progress}
  \end{subfigure}
    \hfill 
  \begin{subfigure}[t]{0.3\columnwidth}
    \includegraphics[width=\linewidth]{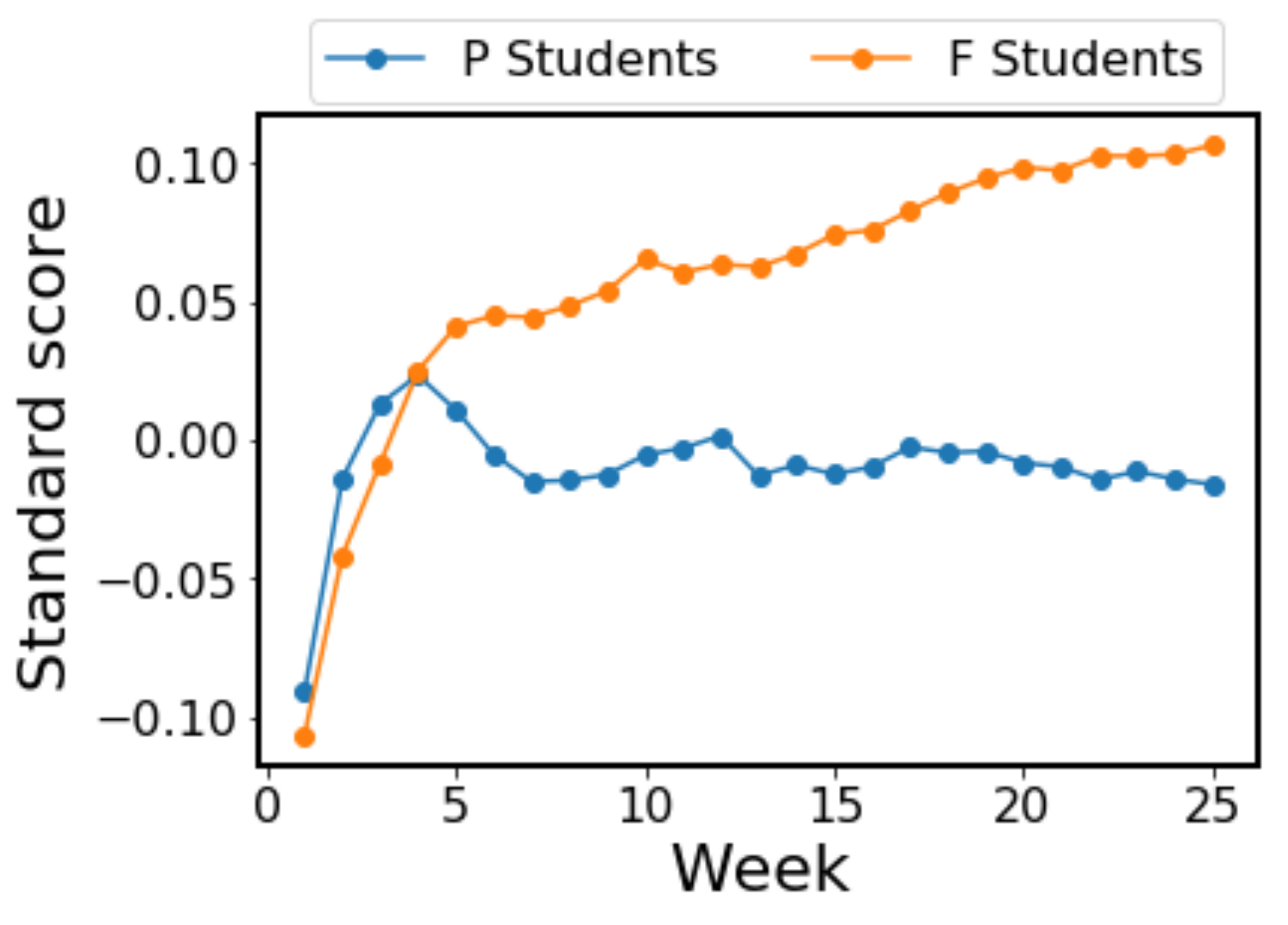}
    \caption{Topic "Assessment"}
    \label{fig:term_progress}
  \end{subfigure}
  
    \begin{subfigure}[t]{0.3\columnwidth}
    \includegraphics[width=\linewidth]{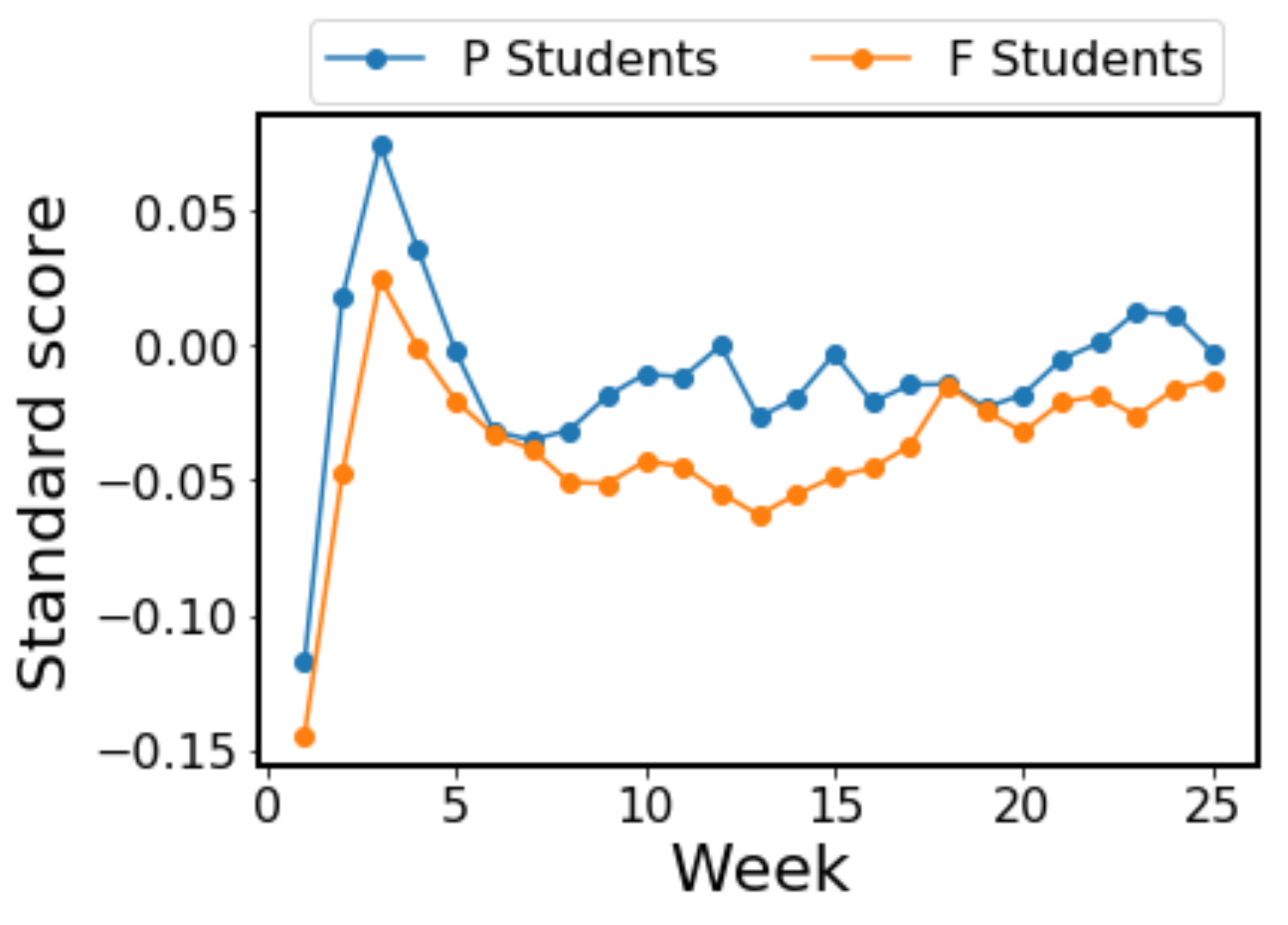}
    \caption{Topic "Review for exam"}
    \label{fig:case_complete}
  \end{subfigure}
  \hfill 
  \begin{subfigure}[t]{0.29\columnwidth}
    \includegraphics[width=\linewidth]{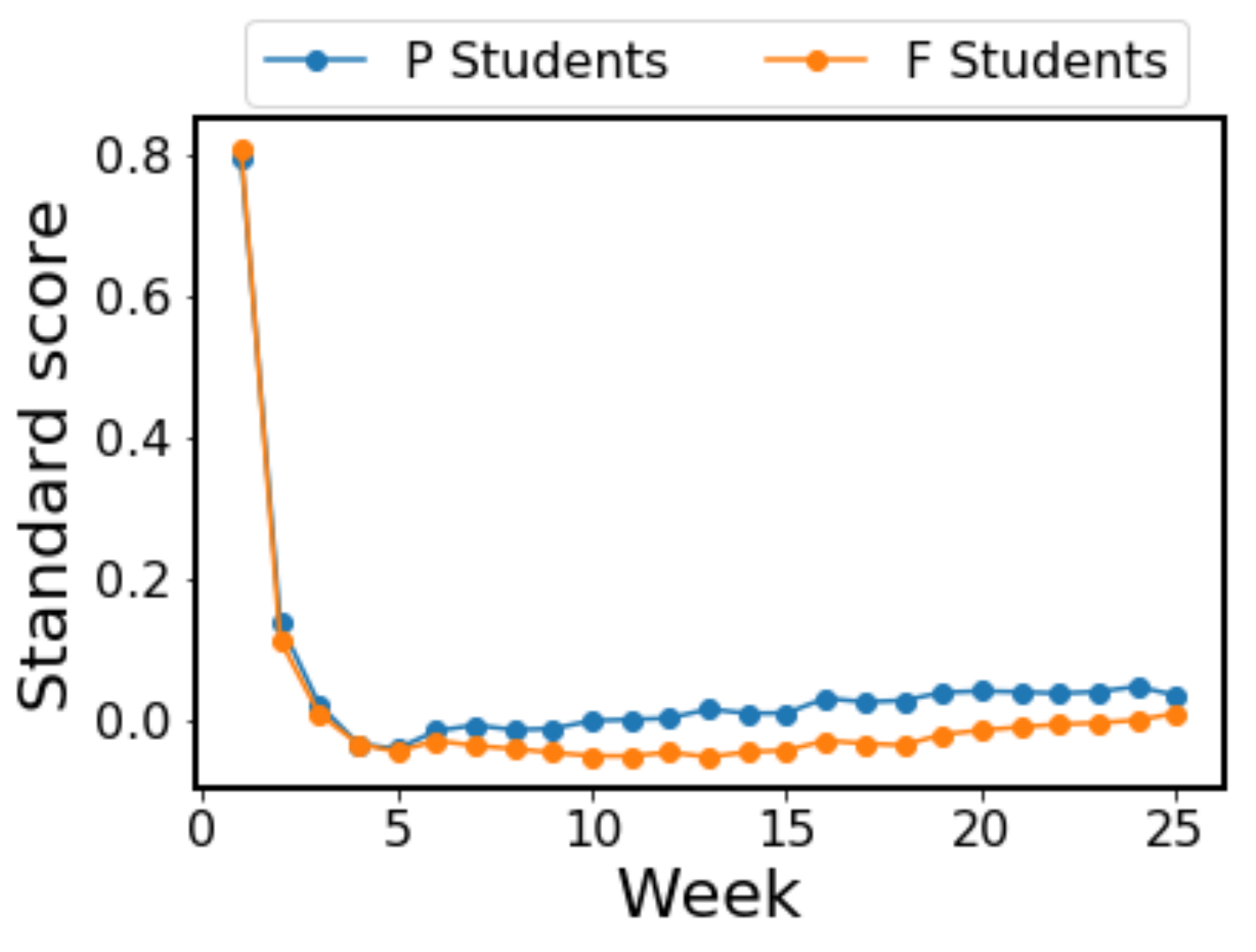}
    \caption{Topic "Term plan"}
    \label{fig:term_progress}
  \end{subfigure}
    \hfill 
  \begin{subfigure}[t]{0.3\columnwidth}
    \includegraphics[width=\linewidth]{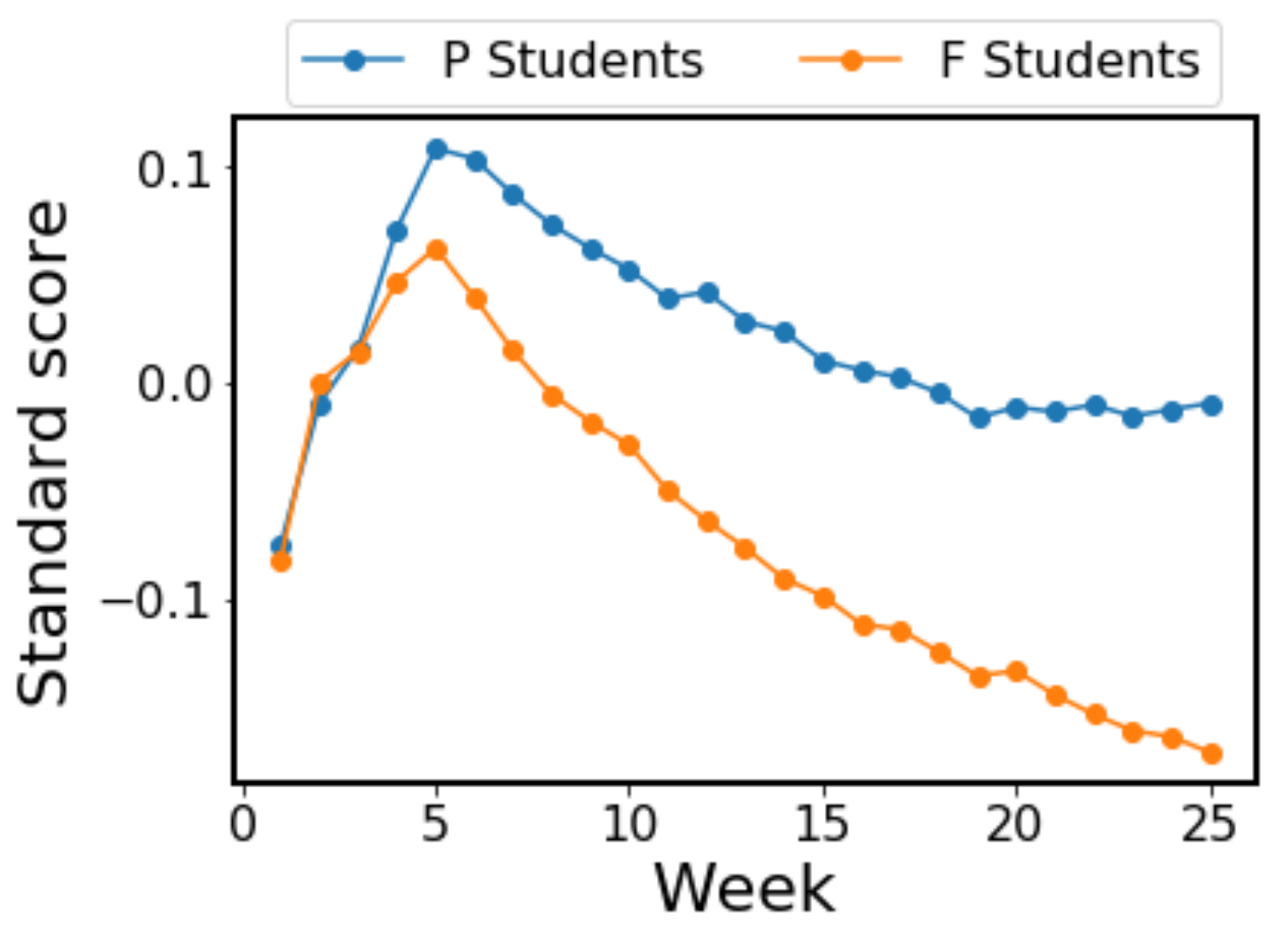}
    \caption{Topic "Course progress"}
    \label{fig:term_progress}
  \end{subfigure}

  \begin{subfigure}[t]{0.3\columnwidth}
    \includegraphics[width=\linewidth]{FIG/case_course_progress.pdf}
    \caption{Topic "Term progress"}
    \label{fig:case_complete}
  \end{subfigure}
%
  \hspace{7mm}
  \begin{subfigure}[t]{0.3\columnwidth}
    \includegraphics[width=\linewidth]{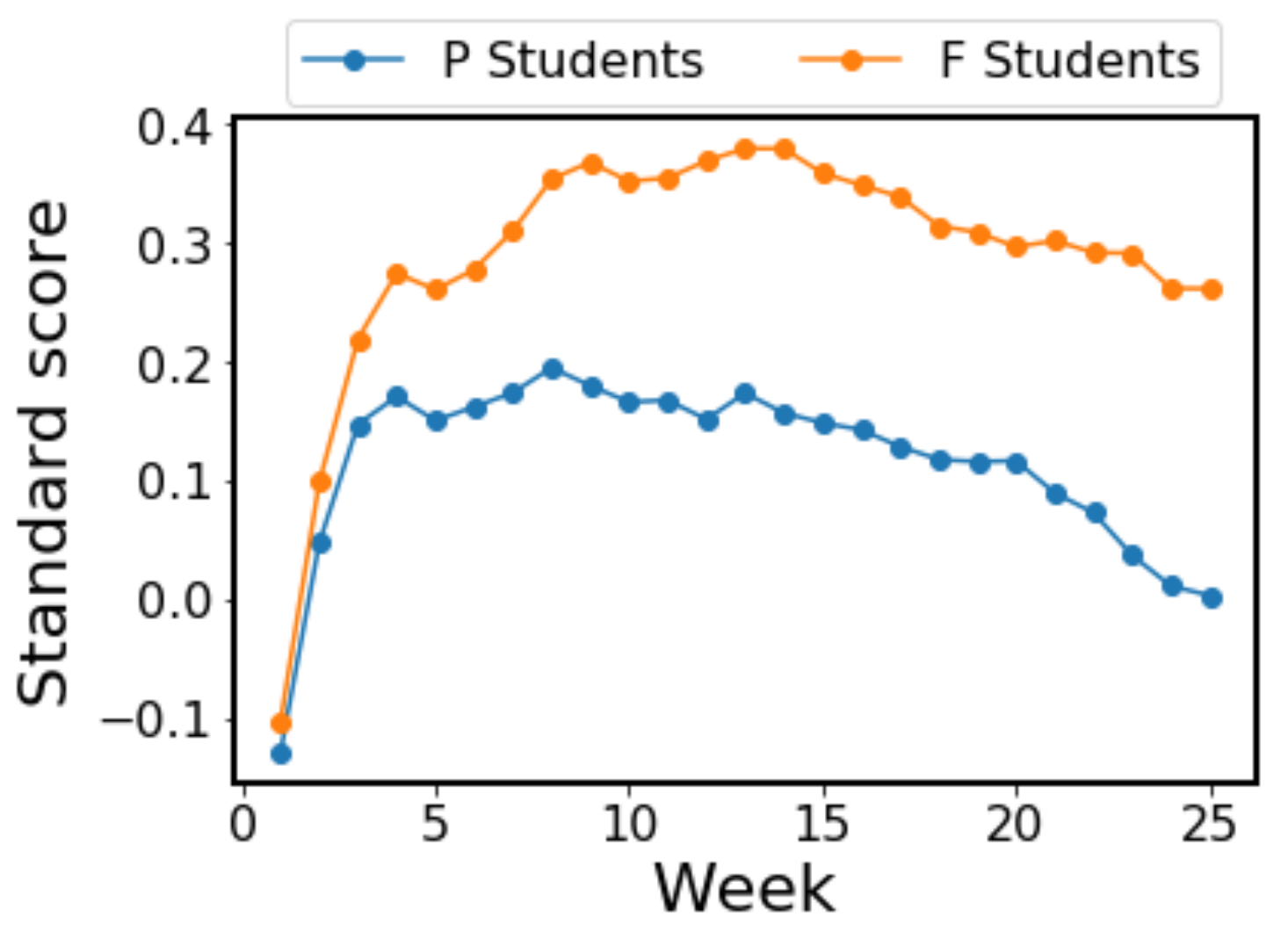}
    \caption{Topic "Time constraints"}
    \label{fig:term_progress}
  \end{subfigure}
  \hspace{6mm}
  \begin{subfigure}[t]{0.3\columnwidth}
    \includegraphics[width=\linewidth]{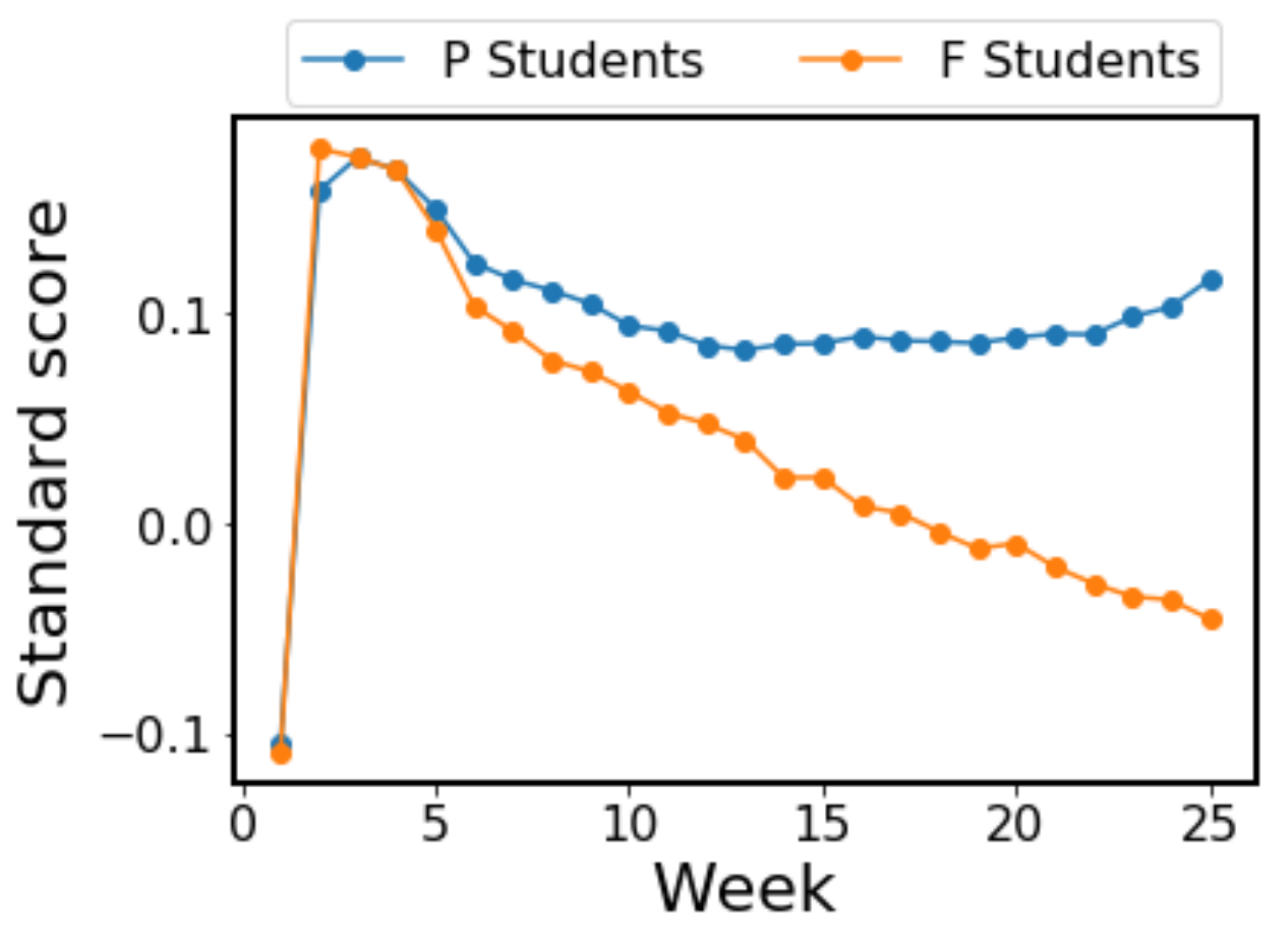}
    \caption{Topic "Goal setting"}
    \label{fig:term_progress}
  \end{subfigure}
  \caption{Standard Score of Each Topic Probability across Weeks for P and F Students}
  \label{fig:student_topics}
\end{figure*}

\textbf{RQ3. What insights do we gain about the process of passing or failing a course over time from predicted mentor's notes topic distributions over time from the model?}  To answer the question of characterizing students likely to pass or fail a course, we perform three different experiments on the dataset of clickstream and mentors' notes data of students taking the College Algebra course. We choose this course because our topic prediction loss was lower (and thus, accuracy higher) for the course. First, we determine what topics inferred from our model correlate with whether a student will pass or fail a course (Experiment 1). Then we find sequences of standardized topic probabilities of each topic inferred by our model that characterize students likely to pass or fail (Experiment 2). Lastly, we report how early we can identify students likely to fail based on our learned student state representation (Experiment 3).

\begin{table}[H]
    \small
    \centering
        \begin{tabularx}{\textwidth}{c|c|c|c}
            \specialrule{.12em}{.1em}{.1em} 
            State & Revision & Question & Assessment\\
            \specialrule{.12em}{.1em}{.1em} 
            P & 0.4795 & 0.0346 & \textbf{-0.1486} \\
            \hline
            F & 0.4914 & 0.1090 & \textbf{0.1735}\\
            \specialrule{.12em}{.1em}{.1em} 
            State & Review for exam & Term plan & Course progress\\
            \specialrule{.12em}{.1em}{.1em} 
            P & -0.1469 & -0.0481 & \textbf{0.6301}  \\
            \hline
            F & -0.0752 & -0.1426 & \textbf{-0.0049} \\
            \specialrule{.12em}{.1em}{.1em} 
            State & Term progress & Time constraint & Goal setting\\
            \specialrule{.12em}{.1em}{.1em} 
            P & \textbf{0.2298} & -0.4811 & 0.0817 \\
            \hline
            F & \textbf{-0.1647} & -0.2186 & 0.0661 \\
            \specialrule{.12em}{.1em}{.1em} 
        \end{tabularx}
    
    \caption{Standard Score of Inferred Topic Probabilities from P and F State}
    \label{tab:fail_or_pass}
\end{table}
\vspace{-2em}

\textbf{Experiment 1.} In the first experiment, we find the two student state representations which minimize or maximize the probability of failing a course. We call the state representations that minimize and maximize the probability of failure as a P state and F state, respectively. Then, we show what topic distributions are inferred from P and F states. We represent emphasis, or a lack thereof, on a topic by standardizing topic probabilities and observing the number of standard deviations above and below the mean of a topic probability.

Table \ref{tab:fail_or_pass} shows the number of standard deviations above or below the mean (also called standard score) for each inferred topic probability from the P and F state. While the standard score of some topics are similar for the P and F state, some vary wildly. For example, the standard score of assessment topic (T3) for the P state is negative and for the F state is positive. One interpretation of this results is that students likely to fail have more trouble in passing assessments, and thus talked to their mentors more about assessment topic (T3). As shown in Table \ref{tab:topic}, the topical text of assessment (T3) includes a general description of students' progress in assessments. The standard score of course progress (T6) and term progress (T7) for the P state is positive and negative for the F state. This suggests that students more likely to pass show smoother progress instead of reporting ongoing issues, so their typical mentors' notes can focus exclusively on progress reports. Evidently, the topical text of course and term progress (T6, T7) in Table \ref{tab:topic} is a progress report without ongoing issues.

\textbf{Experiment 2.} We investigate the inferred probability of each topic from learned student state representation by our model. We compare the trajectory of inferred probability of each topic from students who passed (P students) and failed (F students) a course. Figure \ref{fig:student_topics} shows the average standard score of topic probability per topic for P and F students over time.

We can see through this experiment clear, distinct patterns for the frequency of each topic over time that make intuitive sense given the format of online courses. For example, term plan (T5) is high frequency for the first week and plunges right after, since most students and mentors will naturally discuss plans for a term at the start of each term. The topical text of term plan (T5) actually exhibits the discussion of course choice for the final term, expected to happen at the beginning of the final term. The standard scores of other topics related to goal and progress (T6,T7,T9) also decrease over time, likely for similar reasons. The standard scores of revision (T1), question (T2), and assessment (T3), meanwhile, increase over time, which may indicate students seek help more actively as they approach the end of a term. The standard scores of review for exam (T4) increase dramatically until the third week, decrease for few weeks, and finally level off. As the only condition for students in WGU to pass a course is to pass the final assessment, it may be that many students take their final assessments during the earlier weeks so they can pass a course as early as possible. The standard scores of time constraints (T8) steeply increase until the fourth week, and then gradually decrease over time. This suggests that when students begin a term they do not expect to have time constraints, but accumulate unanticipated issues in their personal lives as the course goes on. It is possible that students mention time constraints less near the end of the term either because they have already abandoned the course or they better planned out their schedules near the end of the term in order to complete a term. However, we need to keep in mind that it may not be that they actually have time constraints, but that they just needed an excuse for slow progress. The topical text in Table \ref{tab:topic} also shows time constraints may be discussed as the potential reason for slow progress though it may not be the real reason for the issue.  

For most topics, the P and F students exhibit distinct divergences in topic patterns. For topics related to goal and progress (T6, T7, T9), the gap between P students and F students increases over time--suggesting that as time goes on F students will be reporting issues and obstacles to their mentors instead of positive progress. The gap between P students and F students for question (T2) increases over time, likely for similar reasons. For revision (T1), P students generally have higher standard scores than F students over time, which supports the idea that P students actively seek opportunities for revision and review towards the end of a term. For assessment (T3), standard score for F students increases over time while score for P students decreases. This could suggest that F students are more likely to procrastinate and struggle with their assessments than P students. Finally, for time constraints (T8) F students show higher standard score as time goes on. A likely interpretation is that students who encounter time constraints due to a demanding job or other outstanding personal issues cannot devote focus to a course and are more likely to fail.



\textbf{Experiment 3.} We report how early we can identify students likely to fail based on our learned student state representation. We compare the Euclidean distance between student state representation and the representation of P and F states. Then we plot the average distance of students who actually passed and failed (P student and F student) across time steps.

\begin{figure}[H]
  \begin{subfigure}[b]{0.64\columnwidth}
    \includegraphics[width=\linewidth]{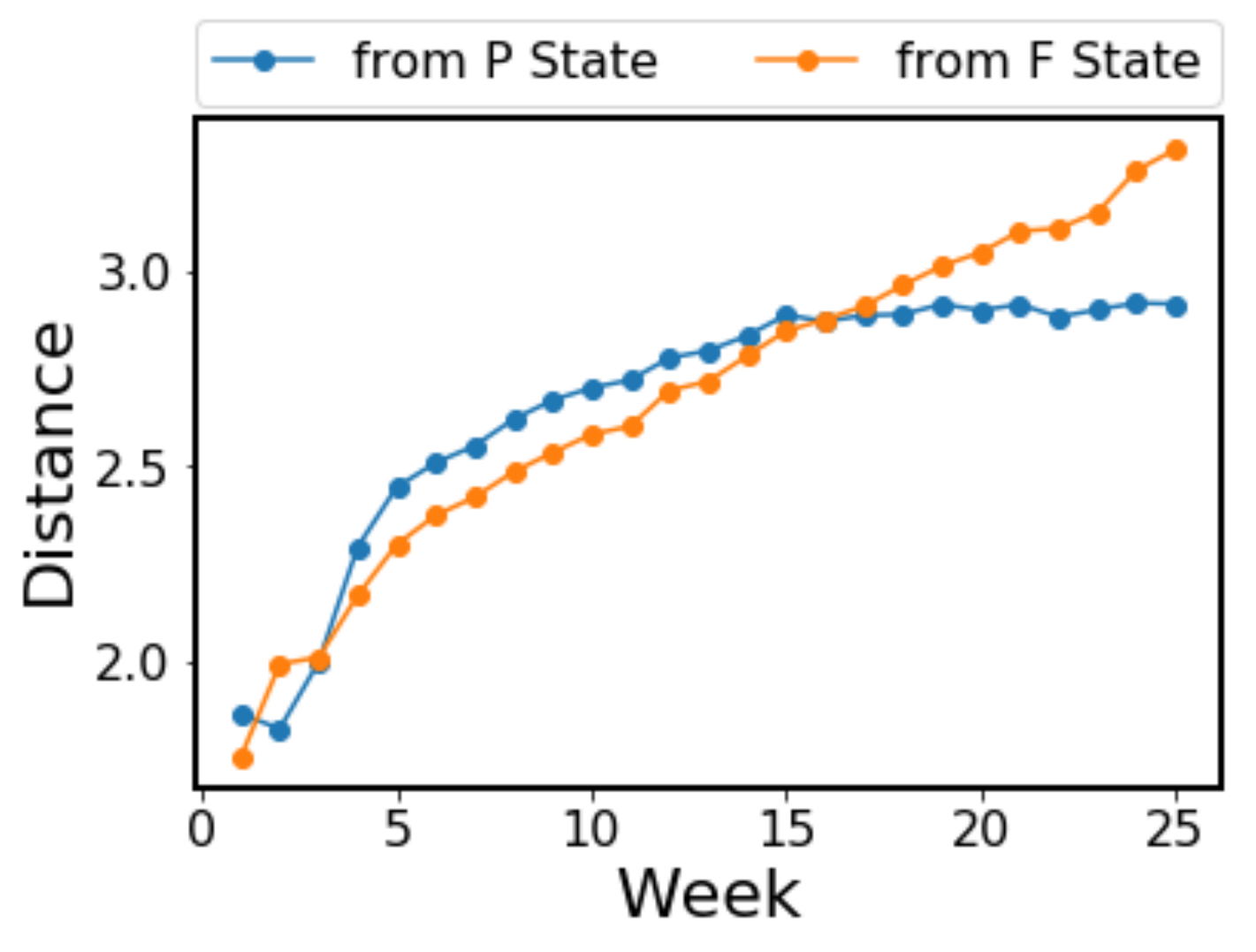}
    \caption{P Students}
    \label{fig:state_p}
  \end{subfigure}
  \begin{subfigure}[b]{0.64\columnwidth}
    \includegraphics[width=\linewidth]{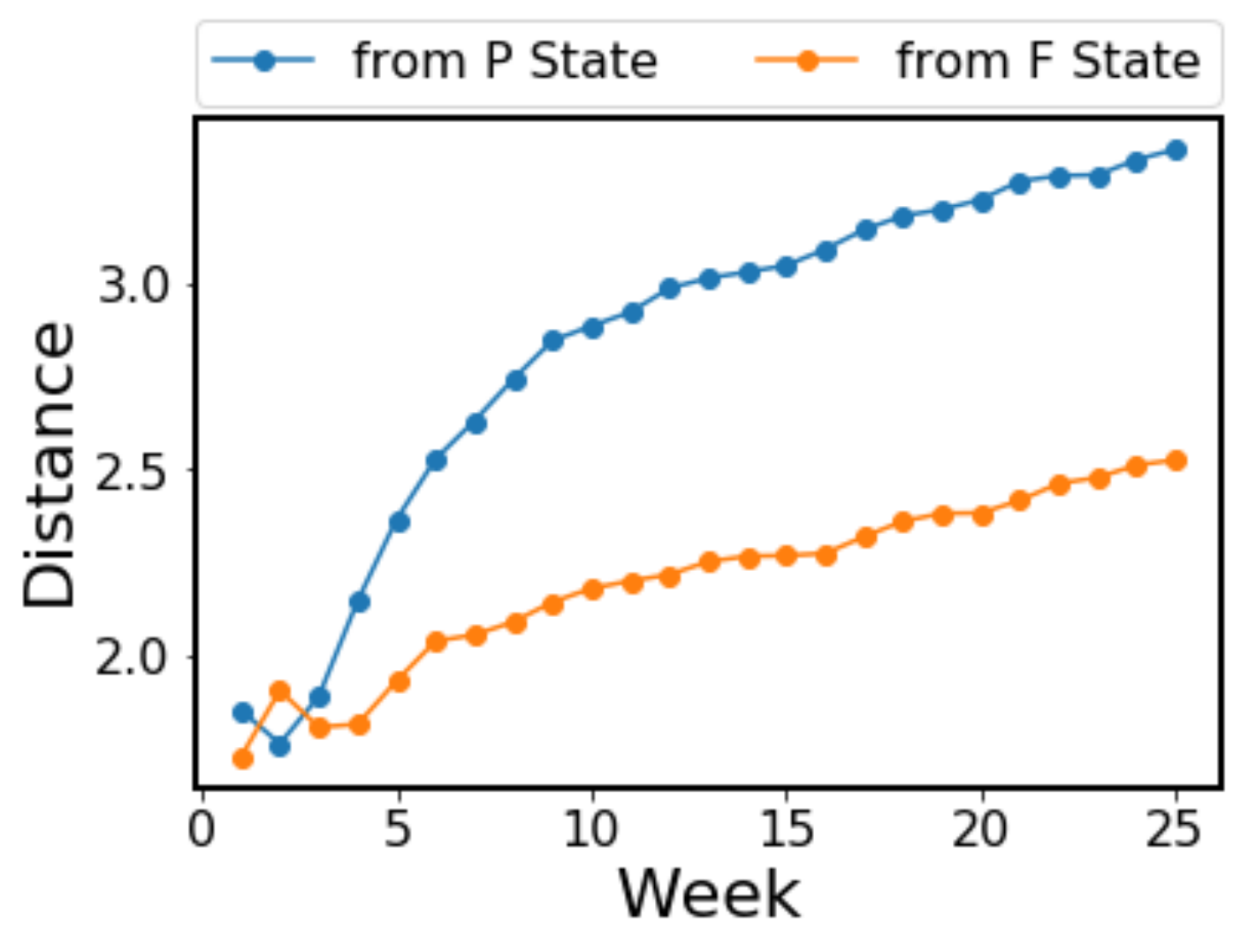}
    \caption{F Students}
    \label{fig:state_f}
  \end{subfigure}
  \caption{Average Euclidean Distance between State Representation of P and F Students and P and F State Representation}
  \label{fig:distance}
\end{figure}

Figure \ref{fig:distance} shows how the Euclidean distance between the student state and the P and F states changes over time. We find that F students are closer to the F state over all different weeks except second week. P students on average are approximately the same distance between the P and F state until about 15 weeks. Based on these observations, we can see in early stages whether a student is likely to fail from the difference between the distance of our model's student state representation from P and F states. This is especially evident in early stages (week 5 onward), where we see a substantial distance between the average F student's state representation from the P state compared to the F state, while there is much more ambiguity for P students.

%% file: 060conclusion.tex
In this paper, we propose and evaluate a sequence model, \textbf{Click2State}, which aims to build an interpretable student state representation by leveraging mentor's notes to give deeper meaning to impoverished clickstream data. We also introduce a methodology for interpreting the learned representation from our model that extracts time-sensitive insights about the process of passing or failing a given online course. Our experimental results demonstrate that student state representations learned by our model have better predictive power on the task of determining student failure rate than a baseline that only uses click stream data. We also present how individual topic-based insights into the process of passing or failing a course let us construct a rich characterization of a student likely to fail or pass an online course.

We see many possible ways to build on the work in this paper. For instance, instead of learning topics using a simple LDA approach, which is purely unsupervised, we might train our model to learn topics that directly correlate with a student's likelihood of failure \citep{mcauliffe2008supervised,  DBLP:journals/corr/DiengWGP16}. In this case, the key challenge might be to train a model in a way that enable us to maintain the level of interpretability in topics that we see from LDA. Another direction could utilize mentor's notes explicitly as an additional input of a model. This approach might provide us a more targetted student state representation.

%% file: working_note.bbl
\begin{thebibliography}{10}

\bibitem{ashby2004monitoring}
Alison Ashby.
\newblock Monitoring student retention in the open university: definition,
  measurement, interpretation and action.
\newblock {\em Open Learning: The Journal of Open, Distance and e-Learning},
  19(1):65--77, 2004.

\bibitem{blei2003latent}
David~M Blei, Andrew~Y Ng, and Michael~I Jordan.
\newblock Latent dirichlet allocation.
\newblock {\em Journal of machine Learning research}, 3(Jan):993--1022, 2003.

\bibitem{bosch2014s}
Nigel Bosch, Yuxuan Chen, and Sidney D'Mello.
\newblock It’s written on your face: detecting affective states from facial
  expressions while learning computer programming.
\newblock In {\em International Conference on Intelligent Tutoring Systems},
  pages 39--44. Springer, 2014.

\bibitem{botelho2017improving}
Anthony~F Botelho, Ryan~S Baker, and Neil~T Heffernan.
\newblock Improving sensor-free affect detection using deep learning.
\newblock In {\em International Conference on Artificial Intelligence in
  Education}, pages 40--51. Springer, 2017.

\bibitem{Botelho2018StudyingAD}
Anthony~F. Botelho, Ryan Shaun~Joazeiro de~Baker, Jaclyn Ocumpaugh, and Neil~T.
  Heffernan.
\newblock Studying affect dynamics and chronometry using sensor-free detectors.
\newblock In {\em EDM}, 2018.

\bibitem{calvert2014developing}
Carol~Elaine Calvert.
\newblock Developing a model and applications for probabilities of student
  success: a case study of predictive analytics.
\newblock {\em Open Learning: The Journal of Open, Distance and e-Learning},
  29(2):160--173, 2014.

\bibitem{calvo2010affect}
Rafael~A Calvo and Sidney D'Mello.
\newblock Affect detection: An interdisciplinary review of models, methods, and
  their applications.
\newblock {\em IEEE Transactions on affective computing}, 1(1):18--37, 2010.

\bibitem{cho2014properties}
Kyunghyun Cho, Bart Van~Merri{\"e}nboer, Dzmitry Bahdanau, and Yoshua Bengio.
\newblock On the properties of neural machine translation: Encoder-decoder
  approaches.
\newblock {\em arXiv preprint arXiv:1409.1259}, 2014.

\bibitem{craig2004affect}
Scotty Craig, Arthur Graesser, Jeremiah Sullins, and Barry Gholson.
\newblock Affect and learning: an exploratory look into the role of affect in
  learning with autotutor.
\newblock {\em Journal of educational media}, 29(3):241--250, 2004.

\bibitem{DBLP:journals/corr/DiengWGP16}
Adji~B. Dieng, Chong Wang, Jianfeng Gao, and John~William Paisley.
\newblock Topicrnn: {A} recurrent neural network with long-range semantic
  dependency.
\newblock {\em CoRR}, abs/1611.01702, 2016.

\bibitem{d2007mind}
Sidney D'mello and Arthur Graesser.
\newblock Mind and body: Dialogue and posture for affect detection in learning
  environments.
\newblock {\em Frontiers in Artificial Intelligence and Applications}, 158:161,
  2007.

\bibitem{eagle2018wgu}
Michael Eagle, Ted Carmichael, Jessica Stokes, Mary~Jean Blink, John Stamper,
  and Jason Levin.
\newblock Predictive student modeling for interventions in online classes.
\newblock In {\em EDM}, 2018.

\bibitem{fei2015temporal}
Mi~Fei and Dit-Yan Yeung.
\newblock Temporal models for predicting student dropout in massive open online
  courses.
\newblock In {\em ICDMW 2015}, 2015.

\bibitem{heffernan2014assistments}
Neil~T Heffernan and Cristina~Lindquist Heffernan.
\newblock The assistments ecosystem: Building a platform that brings scientists
  and teachers together for minimally invasive research on human learning and
  teaching.
\newblock {\em International Journal of Artificial Intelligence in Education},
  24(4):470--497, 2014.

\bibitem{hochreiter1997long}
Sepp Hochreiter and J{\"u}rgen Schmidhuber.
\newblock Long short-term memory.
\newblock {\em Neural computation}, 1997.

\bibitem{johnstone2005competency}
Douglas Johnstone.
\newblock A competency alternative: Western governors university.
\newblock {\em Change: The Magazine of Higher Learning}, 37(4):24, 2005.

\bibitem{kingma2014adam}
Diederik~P Kingma and Jimmy Ba.
\newblock Adam: A method for stochastic optimization.
\newblock {\em arXiv preprint arXiv:1412.6980}, 2014.

\bibitem{kinser2002taking}
Kevin Kinser.
\newblock Taking wgu seriously: Implications of the western governors
  university.
\newblock {\em Innovative Higher Education}, 26(3):161--173, 2002.

\bibitem{krizhevsky2012imagenet}
Alex Krizhevsky, Ilya Sutskever, and Geoffrey~E Hinton.
\newblock Imagenet classification with deep convolutional neural networks.
\newblock In {\em Advances in neural information processing systems}, pages
  1097--1105, 2012.

\bibitem{liu2005empirical}
Changchun Liu, Pramila Rani, and Nilanjan Sarkar.
\newblock An empirical study of machine learning techniques for affect
  recognition in human-robot interaction.
\newblock In {\em 2005 IEEE/RSJ International Conference on Intelligent Robots
  and Systems}, pages 2662--2667. IEEE, 2005.

\bibitem{matuga2009self}
Julia~M Matuga.
\newblock Self-regulation, goal orientation, and academic achievement of
  secondary students in online university courses.
\newblock {\em Journal of Educational Technology \& Society}, 12(3):4, 2009.

\bibitem{mcauliffe2008supervised}
Jon~D Mcauliffe and David~M Blei.
\newblock Supervised topic models.
\newblock In {\em Advances in neural information processing systems}, pages
  121--128, 2008.

\bibitem{nicolaou2011continuous}
Mihalis~A Nicolaou, Hatice Gunes, and Maja Pantic.
\newblock Continuous prediction of spontaneous affect from multiple cues and
  modalities in valence-arousal space.
\newblock {\em IEEE Transactions on Affective Computing}, 2011.

\bibitem{pardos2014affective}
Zachary~A Pardos, Ryan~SJD Baker, Maria~OCZ San~Pedro, Sujith~M Gowda, and
  Supreeth~M Gowda.
\newblock Affective states and state tests: Investigating how affect and
  engagement during the school year predict end-of-year learning outcomes.
\newblock {\em Journal of Learning Analytics}, 1(1):107--128, 2014.

\bibitem{rienties2016analytics4action}
Bart Rienties, Avinash Boroowa, Simon Cross, Chris Kubiak, Kevin Mayles, and
  Sam Murphy.
\newblock Analytics4action evaluation framework: A review of evidence-based
  learning analytics interventions at the open university uk.
\newblock {\em Journal of Interactive Media in Education}, 2016(1), 2016.

\bibitem{rodrigo2009affective}
Ma~Mercedes~T Rodrigo, Ryan~S Baker, Matthew~C Jadud, Anna Christine~M Amarra,
  Thomas Dy, Maria Beatriz~V Espejo-Lahoz, Sheryl Ann~L Lim, Sheila~AMS Pascua,
  Jessica~O Sugay, and Emily~S Tabanao.
\newblock Affective and behavioral predictors of novice programmer achievement.
\newblock In {\em ACM SIGCSE Bulletin}. ACM, 2009.

\bibitem{SinhaJLD14}
Tanmay Sinha, Patrick Jermann, Nan Li, and Pierre Dillenbourg.
\newblock Your click decides your fate: Leveraging clickstream patterns in
  {MOOC} videos to infer students' information processing and attrition
  behavior.
\newblock {\em CoRR}, abs/1407.7131, 2014.

\bibitem{Sinha2014}
Tanmay Sinha, Nan Li, Patrick Jermann, and Pierre Dillenbourg.
\newblock Capturing "attrition intensifying" structural traits from didactic
  interaction sequences of mooc learners.
\newblock In {\em EMNLP 2014, Workshop on Analysis of Large Scale Social
  Interaction in MOOCs}, 2014.

\bibitem{smith2012predictive}
Vernon~C Smith, Adam Lange, and Daniel~R Huston.
\newblock Predictive modeling to forecast student outcomes and drive effective
  interventions in online community college courses.
\newblock {\em Journal of Asynchronous Learning Networks}, 2012.

\bibitem{tang2016deep}
Steven Tang, Joshua~C Peterson, and Zachary~A Pardos.
\newblock Deep neural networks and how they apply to sequential education data.
\newblock In {\em Proceedings of the Third (2016) ACM Conference on Learning@
  Scale}, pages 321--324. ACM, 2016.

\bibitem{DBLP:conf/iccse/WangYM17}
Wei Wang, Han Yu, and Chunyan Miao.
\newblock Deep model for dropout prediction in moocs.
\newblock In {\em ICCSE 2017}, 2017.

\end{thebibliography}
